\documentclass[11pt]{article}
\usepackage[margin=1in]{geometry}
\usepackage{amsmath,amssymb,amsthm,booktabs,array,hyperref,graphicx,natbib}
\newtheorem{theorem}{Theorem}\newtheorem{proposition}{Proposition}\newtheorem{lemma}{Lemma}\newtheorem{corollary}{Corollary}
\theoremstyle{definition}\newtheorem{definition}{Definition}\newtheorem{remark}{Remark}\newtheorem{example}{Example}
\newcommand{\E}{\mathbb{E}}
\title{Verifier Noise Is a Temperature:\\ \large Testing Tilt Predictions in Finite-Budget LLM-Guided Search}
\author{Xiaomeng Li\\ {\normalsize Link Heart Limited, Houston, TX, USA}}
\date{Preprint, November 2026}
\begin{document}
\maketitle
\begin{abstract}
LLM-guided search loops propose a modified candidate, score it, and keep it if it scores better. When the score is noisy and both candidates are re-scored at every comparison, as in the minibatch acceptance gates of prompt optimizers, keep-if-better is a reversible Markov chain if the score difference is logistic and the mutation operator is reversible, and its stationary law is the mutation operator's equilibrium tilted by $e^{\beta f}$. For any continuous symmetric noise difference the same tilt is the stationary law of the small-step diffusion limit, with $\beta=4F'(0)$, where $F$ is the distribution of the noise difference; for a minibatch gate $\beta$ grows with the square root of the minibatch size. Caching the incumbent's score instead inflates the recorded score, and whether iteration beats independent best-of-$N$ depends on the noise and quality tails. We test the tilt as a prediction for LLM-guided program search, committing each prediction from calibration data alone before the search runs. The level after 150 steps rises with the minibatch size in six of seven settings, but its predicted value fails: on bin packing, search attains $62$--$88\%$ of the predicted gain when the proposer edits the parent and $109$--$119\%$ when it writes new heuristics. The attained fraction falls with the parent--child quality correlation of unselected mutation; a registered test on two new settings confirmed the direction, while the quantitative agreement, for the small-edit setting, did not survive a correction for proposals shared across conditions. Two further registered tests found that runs started from a good program kept losing quality over 300 steps and, in the setting tested, that the loss was larger from starts that search had selected than from starts of the same quality visited by unselected walks: after 300 steps the latter were statistically compatible with the prediction and the former narrowly below it, a difference that passed its registered rule by a small margin. Within our budgets, runs from different starts had not reached a common level, so whether the tilt describes the stationary law of LLM editing remains open; on bin packing, for editing proposers, it was an upper estimate of what search from the starting program reached. A registered test in a second domain, prompt optimisation on frozen task-model scores, failed: search exceeded the tilt, and the bin-packing relation between correlation and attained fraction did not transfer.
\end{abstract}

\section{Introduction}
A growing family of systems improves an artifact by letting a language model propose edits and keeping the edits that score better: program search in FunSearch and AlphaEvolve \citep{romera2024,novikov2025}, prompt and pipeline optimisers such as GEPA and MIPRO \citep{agrawal2025gepa,opsahl2024}, and self-refinement loops judged by another model. The score that decides what is kept is usually noisy: a minibatch of tasks, a sampled workload, or a stochastic LLM judge. Practitioners respond to this noise in two ways. Some re-score the incumbent together with each candidate on a fresh sample, as in GEPA's minibatch acceptance gate. Others score each candidate once and store the result, as in program databases.

This paper characterises the first choice exactly and contrasts it with the second. Our starting observation is classical: when both candidates are re-scored with independent Gumbel noise, keep-if-better is Barker's acceptance rule, and the search is a reversible Markov chain whose stationary law is the mutation operator's equilibrium tilted by $e^{f/b}$. This is the logit dynamics of game theory and the mutation--selection balance of population genetics with ``noise scale'' in place of ``population size''; that payoff noise raises the temperature of selection is known for evolutionary games \citep{traulsen2007stochastic}. We take it as a lemma and derive what it implies for LLM-guided search:
\begin{enumerate}
\item \textbf{Any symmetric noise is a temperature in a small-step diffusion limit.} The limiting diffusion has the tilted law as its stationary law, with inverse temperature $\beta=4F'(0)$, four times the density of the score-noise difference at zero (Theorem~\ref{thm:smallstep}); for Gaussian noise of standard deviation $\sigma$ per evaluation, $\beta=2/(\sqrt\pi\,\sigma)$, and for a minibatch gate $\beta\propto\sqrt m$.
\item \textbf{Under these conditions, re-evaluated search has a computable stationary level,} $f^\star=\kappa'(\beta)$, the derivative of the cumulant generating function of quality under the mutation operator's own equilibrium (Corollary~\ref{cor:plateau}). The verifier enters only through $\beta$ and the LLM only through the distribution of quality that unselected mutation produces, which can be measured. Whether a finite run reaches this level is a separate, empirical question.
\item \textbf{Caching the score inflates the reported score} and changes the dynamics; whether true quality keeps improving depends on the mutation operator (with state-independent mutation, caching is best-of-$N$ and also plateaus). In simulation, cached search is ahead of re-evaluation at every budget. In an exploratory LLM run it is not ahead within 400 evaluations, while its recorded score overstates true quality by up to three times the gain that a larger minibatch buys.
\item \textbf{The noise tail, together with the quality tail, decides} whether iterative search can beat independent best-of-$N$ at matched evaluations: with Gaussian noise and unbounded Gaussian quality, best-of-$N$ eventually wins and re-evaluated search leads only in a window; under Gumbel noise both converge to tilts of different base laws (Section~4).
\end{enumerate}
We check these predictions in simulation, in LLM-guided search over online bin-packing and travelling-salesman heuristics and, in one registered test, over prompts, in all of which the noise is a random minibatch of instances and the true quality is the score on all instances (Section~\ref{sec:llm}). Each prediction uses only a calibration run of the mutation operator without selection, and is committed before the search is launched. Across seven settings (two tasks; two proposer models; three mutation instructions) the level after 150 steps rises with the minibatch size in six, which any account of selection under noise predicts; the predicted value is what discriminates, and it fails. On bin packing it is too high by $0.0008$--$0.0024$ when the proposer edits the parent, and too low when it writes new heuristics whose quality barely depends on the parent. The error ordered with the parent--child quality correlation of unselected mutation, and a registered test on two new settings at the two ends of the correlation range confirmed the direction; the quantitative agreement passed its registered rule in both settings but, in the small-edit setting, not a correction for proposals shared across minibatch sizes, and with the uncertainty of the new settings' calibration and of the fit approximately propagated it is neither established nor refuted. A third registered test continued four editing settings to 300 steps: with the small-edit instruction, runs started from one high-quality program kept losing quality and met or crossed the runs from below under the prediction, while with the neutral instruction slow mixing was not clearly rejected. Because that start had itself been selected by a search, a fourth registered test started runs from twelve programs of matched quality, half selected by search and half visited by unselected walks. It was inconclusive on its primary window (the first 50 steps); on its secondary window (steps 100--150) the selected starts lost $0.0019$ and the walk-visited starts $0.0003$, in the one registered setting, and a registered continuation found the two groups still apart after 300 steps, by a margin that passed the registered rule narrowly and does not survive leaving out most single starts, with the walk-visited starts statistically compatible with the prediction and the selected ones just below it. Among the tested starts, whose origin was not randomised, the quality loss of runs from above is thus associated with where the start came from, while single-step improvement rates show no robust difference once a sampling bias in their estimate is removed and errors are clustered by run. A fifth registered test took the inheritance relation to a second domain, minibatch-gated prompt optimisation on frozen task-model scores, and it failed there: in both settings search exceeded the tilt, the setting with the higher correlation by more, opposite to bin packing; one of the two settings met the transferred line, which, with calibration uncertainty included, means only that the line was not rejected there. Table~\ref{tab:hyp} lists every registered hypothesis, its verdict and the robustness checks. Within our budgets runs from different starts had not reached a common level, so we cannot say whether the tilt is the stationary law of LLM editing; what the data support is narrower: an effective-temperature proxy for the gate can be computed from calibration, on bin packing the tilt is an upper estimate of the level that editing proposers reached from the starting program in a few hundred steps, and there the inheritance correlation, measured without selection, indicated how far below or above it the runs ended; neither statement carried over to the prompt domain.

\section{Setting}
Let $\mathcal X$ be a space of candidates (programs, prompts, plans) with true quality $f:\mathcal X\to\mathbb R$, the quantity one would measure with unlimited evaluation. A \emph{mutation operator}---in practice an LLM prompted with the incumbent---is a Markov kernel $K(x,\mathrm dy)$. We write $\pi_0$ for its equilibrium: the distribution that repeated mutation \emph{without selection} converges to.

\begin{definition}[Re-evaluated keep-if-better search, regime R]
Given incumbent $x_t$, draw $y\sim K(x_t,\cdot)$, score both on fresh noise, $\hat s(x_t)=f(x_t)+\varepsilon$, $\hat s(y)=f(y)+\varepsilon'$, and set $x_{t+1}=y$ if $\hat s(y)>\hat s(x_t)$, else $x_{t+1}=x_t$.
\end{definition}
With $D=\varepsilon-\varepsilon'$ and $\Delta=f(y)-f(x)$, the acceptance probability is $a(x,y)=\Pr(D<\Delta)=F_{xy}(\Delta^-)$, the left limit at $\Delta$ of the distribution function $F_{xy}$ of $D$; we write $F_{xy}(\Delta)$ where $F_{xy}$ is continuous. If the two noise draws are exchangeable and $D$ has a continuous law, $F_{xy}(\Delta)+F_{xy}(-\Delta)=1$. This covers i.i.d.\ continuous noise. It does not cover paired minibatch gates exactly, where parent and child are scored on the same freshly sampled minibatch of $m$ tasks: there $-D$ is the minibatch mean of the per-task score differences minus its population mean, whose law is that of the actual differences; it need not be symmetric for small $m$ (with child-minus-parent differences $-1,-1,2$ on three tasks and $m=1$, the child wins with probability $1/3$ although $\Delta=0$), and it has atoms when scores are discrete. For large $m$ it is approximately Gaussian with standard deviation $\tau_{xy}/\sqrt m$, and the symmetric formulas below apply to it as an approximation (Section~3.2).

\paragraph{Regime C (cached).} The incumbent keeps the score it was accepted with, and a proposal is accepted if its single fresh score exceeds that cached value. This is how program databases in FunSearch-style systems operate. The comparison is no longer symmetric: the cached score is the maximum of the noisy scores the chain has accepted, so it is biased upward (the winner's curse).

\paragraph{Independent sampling (IID).} Draw $N$ candidates from a fixed generator $\pi_{\mathrm{root}}$ (fresh samples from the seed prompt), score each once, and return the arg-max of the noisy score.

\paragraph{Budget.} We compare methods at a matched number of \emph{evaluations}: IID uses one per candidate, regime C one per step, regime R two per step.

\paragraph{Assumption (reversibility).} For the results of Section~3 we assume $K$ is reversible with respect to $\pi_0$. This holds for the Gaussian autoregressive kernels used in our simulations, and is the standard modelling assumption under which mutation-selection balance is analysed in population genetics. LLM mutation operators are not reversible in general; Section~5 tests how much the predictions depend on this assumption.

\section{Stationary law of re-evaluated search}\label{sec:theory}
\subsection{Exact tilt under Gumbel noise}
\begin{lemma}[Barker acceptance gives an exponential tilt]\label{lem:barker}
If $a(x,y)=\sigma\big(\beta(f(y)-f(x))\big)$ with $\sigma(u)=1/(1+e^{-u})$, $K$ is $\pi_0$-reversible, and $Z(\beta)=\int e^{\beta f}\,\mathrm d\pi_0<\infty$, then regime R is reversible with respect to
\[\pi_\beta(\mathrm dx)=Z(\beta)^{-1}e^{\beta f(x)}\,\pi_0(\mathrm dx).\]
\end{lemma}
\begin{proof}
For $x\neq y$, $\pi_\beta(x)K(x,y)a(x,y)=Z^{-1}\pi_0(x)K(x,y)\,e^{\beta f(x)}\sigma(\beta\Delta)$ with $\Delta=f(y)-f(x)$. Since $\sigma(\beta\Delta)=e^{\beta\Delta}\sigma(-\beta\Delta)$ and $\pi_0(x)K(x,y)=\pi_0(y)K(y,x)$, this equals $\pi_\beta(y)K(y,x)a(y,x)$.
\end{proof}
The difference of two independent Gumbel variables of scale $b$ is logistic of scale $b$ \citep{mcfadden1974,yellott1977}, so i.i.d.\ Gumbel score noise gives exactly Barker's rule \citep{barker1965} with $\beta=1/b$. The same composition---noise re-drawn at each pairwise comparison, logistic choice, Gibbs stationary law---is classical in logit dynamics \citep{blume1993,alosferrer2010} and, with population size in place of $1/b$, in population genetics \citep{iwasa1988,sella2005}. We state it as a lemma; the contribution of this section is what it implies for LLM-guided search and how far it extends beyond the logistic case.

An exact tilt for every reversible $K$ requires $a(x,y)/a(y,x)=e^{\beta\Delta}$, i.e.\ a logistic noise difference. For other noise, regime R is not in general reversible with respect to any tilt of $\pi_0$, and we need a limit.

\subsection{Small steps: only the slope at zero matters}
\begin{theorem}[Small-step tilt]\label{thm:smallstep}
Let $\mathcal X=\mathbb R^n$, let $K_h$ be $\pi_0$-reversible kernels with step covariance $h^2I+o(h^2)$ and $\E|y-x|^3=O(h^3)$ uniformly for $x$ in compact sets, whose generators satisfy $h^{-2}(K_h-I)g\to\frac12(\Delta g+\nabla\log\pi_0\cdot\nabla g)$ for smooth $g$, let $f$ be smooth with bounded first and second derivatives, and let $a(x,y)=F(f(y)-f(x))$ with $F$ differentiable at $0$ and $F(\Delta)+F(-\Delta)=1$. Then the generator of regime R, rescaled by $h^{-2}$, converges to
\[\mathcal L g=\tfrac14\Big(\Delta g+\nabla\big(\log\pi_0+4F'(0)f\big)\cdot\nabla g\Big),\]
a Langevin diffusion with stationary law $\pi_\beta\propto e^{\beta f}\pi_0$, $\beta=4F'(0)$, whenever $\pi_\beta$ is normalisable. Convergence is of generators on smooth compactly supported $g$; the Gaussian autoregressive kernels of Example~\ref{ex:ar} satisfy the moment conditions locally uniformly, not uniformly on $\mathbb R^n$, since their mean step grows linearly in $x$.
\end{theorem}
\begin{proof}[Proof sketch]
Write $F(\Delta)=\tfrac12+F'(0)\Delta+o(\Delta)$. Then $\E[(g(y)-g(x))a(x,y)]=\tfrac12\E[g(y)-g(x)]+F'(0)\E[(g(y)-g(x))(f(y)-f(x))]+o(h^2)$. The first term is $\tfrac12(K_h-I)g$; the second is $F'(0)\,h^2\,\nabla g\cdot\nabla f+o(h^2)$ because the steps have identity covariance at leading order. Dividing by $h^2$ gives $\mathcal L$. Details are in Appendix~\ref{app:proofs}.
\end{proof}
For Gaussian score noise with standard deviation $\sigma$ on each evaluation, $F(\Delta)=\Phi\big(\Delta/(\sqrt2\,\sigma)\big)$ and
\[\beta=\frac{2}{\sqrt\pi\,\sigma}\approx1.128/\sigma.\]
For logistic noise, Theorem~\ref{thm:smallstep} recovers Lemma~\ref{lem:barker}, and the two uses of $\beta$ coincide. The theorem is the stationary-law counterpart of the \emph{universality of weak selection} of \citet{wu2010}, who show that fixation probabilities of pairwise-comparison processes depend on the comparison function only through its slope at zero. The often-used probit-to-logit constant ($1.702/\sqrt2\approx1.2035$ per unit SNR) is the best global logistic fit, not the small-step limit; the two differ by about 7\%.

\paragraph{Minibatch gates.} If the noise comes from scoring parent and child on a fresh minibatch of $m$ tasks with per-task difference standard deviation $\tau$, then $\sigma_{\text{diff}}=\tau/\sqrt m$ and, since $F$ now refers to the \emph{difference},
\[\beta(m)=\frac{4}{\sqrt{2\pi}}\cdot\frac{\sqrt m}{\tau}\approx 1.596\,\frac{\sqrt m}{\tau}.\]
The minibatch size is a temperature knob: $\beta\propto\sqrt m$. If the minibatch is drawn without replacement from a fixed pool of $N$ tasks, the difference has s.d.\ $(\tau/\sqrt m)\sqrt{(N-m)/(N-1)}$, and $\beta(m)$ is multiplied by $\sqrt{(N-1)/(N-m)}$. With discrete scores, a strict gate rejects ties in both directions, so $a(x,y)+a(y,x)=1-p_{xy}$ with $p_{xy}$ the tie probability. Because $p_{xy}=p_{yx}$, ties cancel in the ratio $a(x,y)/a(y,x)$ that enters detailed balance. If $p_{xy}$ is roughly constant over the pairs the chain visits, they only slow the chain, and $\beta$ is $4G'(0)$ for the law $G$ of the difference conditioned on no tie.

\paragraph{The minibatch formula is an approximation.} It is not a corollary of Theorem~\ref{thm:smallstep} without further assumptions. The theorem uses one acceptance function $F$ for all pairs. A paired gate has a pair-dependent $F_{xy}$: its scale $\tau_{xy}$ varies across parent--child pairs, the minibatch mean of per-task differences need not be symmetric about its mean for small $m$, and discrete scores create ties. Edges with different log-odds slopes are not described by one $\beta$, and averaging $\beta_{xy}$ does not restore a Gibbs stationary law. There is also a scaling issue. If per-task scores vary smoothly with the candidate, both the mean and the s.d.\ of the paired difference are $O(h)$ for steps of size $h$, so the signal-to-noise ratio of a comparison need not vanish as $h\to0$, and the fixed-$F$ expansion does not literally apply. We therefore treat $\beta(m)$ as an approximation whose quality we measure (Section~\ref{sec:llm}): it requires roughly Gaussian minibatch means and a $\tau_{xy}$ that varies little over the pairs the chain visits.

\subsection{The plateau}
\begin{corollary}[Plateau]\label{cor:plateau}
Under the conditions of Lemma~\ref{lem:barker} or Theorem~\ref{thm:smallstep}, and if $\E_{\pi_0}\big[|f|e^{\beta f}\big]<\infty$, the stationary mean quality of regime R (under Theorem~\ref{thm:smallstep}, of its small-step diffusion limit) is
\[f^\star(\beta)=\E_{\pi_\beta}[f]=\kappa'(\beta),\qquad \kappa(\beta)=\log\E_{\pi_0}\big[e^{\beta f}\big],\]
the derivative of the cumulant generating function of quality under the mutation equilibrium. If $\E_{\pi_0}e^{t f}<\infty$ for $t$ in a neighbourhood of $[0,\beta]$, $\kappa$ is analytic there and $f^\star(\beta)=\mu_0+\beta\sigma_0^2+\tfrac{\beta^2}{2}\kappa_3+\dots$ for small $\beta$, with $\mu_0,\sigma_0^2,\kappa_3$ the mean, variance and third cumulant of $f$ under $\pi_0$. If $f$ is Gaussian under $\pi_0$, $f^\star=\mu_0+\beta\sigma_0^2$ exactly.
\end{corollary}
\begin{proof}
$\E_{\pi_\beta}[f]=\E_{\pi_0}[fe^{\beta f}]/\E_{\pi_0}[e^{\beta f}]=\kappa'(\beta)$.
\end{proof}
Two practical readings follow. First, the plateau depends on the mutation operator only through the distribution of quality under its own equilibrium, which can be estimated by running the operator \emph{without selection}. To first order a stronger LLM raises the plateau by raising $\mu_0$ (it drifts toward better candidates on its own) or $\sigma_0^2$ (it explores a wider range), and a better verifier raises it through $\beta$. The cumulant expansion is useful only when $\beta\sigma_0$ is small; in our LLM experiments $\beta\sigma_0$ is between $1.4$ and $27$, on bin packing the Gaussian form $\mu_0+\beta\sigma_0^2$ exceeds the maximum possible score of $1$ at $m=64$ in every setting and from $m=16$ on in four of six, and we test the full tilt $\E_{\pi_0e^{\beta f}}f$ instead. Second, Lemma~\ref{lem:barker} shows that $\pi_\beta$ is invariant, not that the chain converges to it from a given start; that needs more, for instance irreducibility and aperiodicity on a finite space. A proposal that never moves between two classes of candidates keeps the chain in the class it starts in, whose own tilt can lie far below the global one. Once the chain has mixed (which our LLM experiments could not establish within their budgets; Section~\ref{sec:llm}), the marginal mean quality of the \emph{current incumbent} no longer increases with the number of steps: extra steps only resample from $\pi_\beta$. This is not an upper bound from every initial condition (a chain started above $f^\star$ drifts down to it), and it does not constrain an output rule that keeps the best candidate ever visited, whose selection would itself need evaluations.

\begin{example}[Autoregressive mutation with drift]\label{ex:ar}
Let quality itself be the state, $y=\rho x+\delta+\sqrt{1-\rho^2}\,z$ with $z\sim N(0,1)$. This kernel is reversible with respect to $\pi_0=N(\delta/(1-\rho),1)$, so under Gumbel noise $\pi_\beta=N(\delta/(1-\rho)+1/b,\,1)$ exactly, and $f^\star=1/b+\delta/(1-\rho)$: a verifier term plus a mutation term. The correlation $\rho$ of the mutation affects the plateau only through the drift term; it mainly sets how fast the plateau is reached.
\end{example}

\subsection{Large steps: slow mixing}
For non-logistic noise and large steps the small-step constant no longer applies; the stationary law of regime R, where it exists, is then not in general a tilt of $\pi_0$, and in the Gaussian autoregressive settings of Table~\ref{tab:beta} its mean lies above $\kappa'(4F'(0))$. Convergence to it can also become very slow. In the autoregressive example with $\rho=0.5$ and Gaussian noise of s.d.\ $1/3$, the stationary mean computed on a grid is $3.996$ ($=1.332\,d$) for grids reaching $x=20$, $30$ and $40$. Simulated search reaches it only slowly: divided by $d$, the mean of 20{,}000 chains is $1.25$, $1.32$, $1.33$ and $1.33$ after $600$, $2{,}500$, $10{,}000$ and $40{,}000$ steps. At $\rho=0.9$ and the same noise it is flat at $1.22$ from 600 steps on. A numerical warning applies to the grid computation. At high quality the probability of leaving a grid state underflows, so many eigenvalues of the transition matrix are numerically equal to one. An eigen-solver can then return a vector concentrated on these numerically absorbing states and suggest that no stationary law exists. We use power iteration started from the bulk, which avoided this in our computations (its results agree across grid ranges and with simulation); it does not exclude truncation effects or very slow mixing in general. Whether regime R has a stationary law in every large-step, non-logistic setting remains open. Corollary~\ref{cor:plateau} is stated where it exists.

\section{Caching, best-of-$N$, and the role of the noise tail}
\subsection{Caching changes the dynamics and inflates the reported score}
In regime C the acceptance threshold is the incumbent's \emph{recorded} score $s_t$, which only ever increases. The incumbent $x_t$ alone is no longer a Markov chain. The pair $(x_t,s_t)$ is a time-homogeneous Markov chain whose score component is non-decreasing. With noise unbounded above, such as Gumbel or Gaussian noise, and proposals whose true quality stays above some fixed level, the recorded score grows without bound and the pair has no stationary law; with bounded scores it can be absorbed at the largest attainable score. In either case Corollary~\ref{cor:plateau} does not apply. This does \emph{not} imply that true quality keeps improving. If the mutation ignores the current state, $K(x,\cdot)=\pi_0$, regime C is exactly best-of-$N$ over independent draws from $\pi_0$. Under Gumbel noise, and for bounded or Gaussian quality, its expected true quality then converges to the tilt mean $\kappa_0'(1/b)$ (Proposition~\ref{prop:bon}a), a plateau, while the recorded score grows without bound. Whether true quality under caching keeps rising therefore depends on the mutation operator and the noise; what is general is the inflation of the recorded score. Two effects compete. Selection keeps acting, because every accepted proposal has to beat a threshold that rises with the best score seen so far. The threshold, however, is inflated by the noise of the lucky draw that set it \citep{arnold2002,smith2006}, so the gap between recorded score and true quality grows and accepted moves become rarer. In the autoregressive model with $d=1$, the recorded score exceeds the incumbent's true quality by $1.98$, $2.89$ and $3.62$ after $32$, $512$ and $8{,}192$ evaluations. With $d=3$ the gaps are $0.47$, $0.74$ and $0.96$ (\texttt{sim\_inflation.py}). In the autoregressive simulation (Section~\ref{sec:sim}), and within the budgets we test, the true quality of regime C keeps rising, leads regime R throughout and overtakes IID. This is an observation for that model, not a general law; the reported score overstates true quality by an amount that grows with the noise and with the budget. \citet{hu2026winner} report related evidence for prompt optimisation (Section~\ref{sec:related}). A practical asymmetry matters as much as caching itself: a re-evaluated gate can score parent and child on the same tasks, which cancels shared task difficulty, while a cached comparison cannot (Section~\ref{sec:llm-explore}). We do not have a closed-form growth law for regime C and present it as an empirical finding.

\subsection{Independent sampling and the noise tail}
For IID best-of-$N$, the returned quality depends on how the noise tail compares with the quality tail. We state two exact cases and one heuristic one.
\begin{proposition}[Best-of-$N$ under two noise tails]\label{prop:bon}
Let $f\sim\pi_{\mathrm{root}}$ and independent noise $\varepsilon$.
(a) \emph{Gumbel noise of scale $b$:} if $0<Z=\E_{\pi_{\mathrm{root}}}e^{f/b}<\infty$, the returned candidate converges in law, as $N\to\infty$, to $e^{f/b}\pi_{\mathrm{root}}/Z$ (Gumbel-max). If moreover $f$ is bounded, or $f$ is Gaussian under $\pi_{\mathrm{root}}$, the expected returned quality converges, $\E[f]\to\kappa_{\mathrm{root}}'(1/b)$.
(b) \emph{Gaussian noise and Gaussian quality} ($f\sim N(\mu,s^2)$, noise s.d.\ $\sigma$): $\E[f]=\mu+e_N\,s^2/\sqrt{s^2+\sigma^2}$, with $e_N$ the expected maximum of $N$ standard normals; this grows without bound like $\sqrt{2\log N}$.
\end{proposition}
\begin{remark}[Heavy-tailed noise]\label{rem:heavy}
\citet[Thm.~5]{kwa2024} show that if the error is subexponential (in the heavy-tailed sense: the class of distributions whose sum-tail behaves like the maximum's) and quality has a lighter tail, then conditioning on a high noisy score gives no gain: $\E[f\mid f+\varepsilon\ge t]\to\E[f]$ as $t\to\infty$. This is a threshold statement, not a statement about the arg-max of $N$ draws. We expect the analogous best-of-$N$ conclusion to hold, but we do not prove the link, and we do not use it below. Their positive result (Thm.~6) requires a tail-ratio condition on the noise and unbounded quality.
\end{remark}
Part (a) is the Gumbel-max trick \citep{verdun2025soft}. Given the qualities, the returned one has mean $\sum_if_ie^{f_i/b}/\sum_je^{f_j/b}$, which converges almost surely to $\kappa_{\mathrm{root}}'(1/b)$ by the law of large numbers; for bounded $f$, dominated convergence gives the mean. For Gaussian quality $N(\mu,s^2)$, Gaussian integration by parts gives the exact identity $\E[f_{I_N}]=\mu+\beta s^2\big(1-\E\sum_ip_i^2\big)$ with $\beta=1/b$ and $p_i=e^{\beta f_i}/\sum_je^{\beta f_j}$, and $\sum_ip_i^2\to0$ almost surely and in mean, so $\E[f_{I_N}]\to\mu+\beta s^2=\kappa_{\mathrm{root}}'(\beta)$ (Appendix~\ref{app:proofs}). A condition on the upper tail alone does not suffice: with $f=-e^{1/U}$, $U$ uniform on $(0,1)$, every exponential moment $\E e^{cf}$, $c>0$, is finite and the tilt has a finite mean, but the returned quality is at most $\max_if_i$, whose expectation is $-\infty$ for every $N$. Part (b) is the breeder's equation with a reliability factor.

\paragraph{Consequences for iterative versus independent search.} Combine Proposition~\ref{prop:bon} with Corollary~\ref{cor:plateau} at matched evaluations. The comparisons are asymptotic and concern expected quality, so both procedures must converge in mean. For IID this holds under the conditions of Proposition~\ref{prop:bon} (bounded or Gaussian quality in part (a)). For regime R we assume that the chain converges in law to its stationary law from the start considered (Lemma~\ref{lem:barker} gives invariance only; on a finite space irreducibility and aperiodicity suffice) and that $f$ is bounded or otherwise uniformly integrable along the chain. The outcome depends jointly on the noise tail, the quality tail and the mutation equilibrium, not on the noise tail alone.
\begin{itemize}
\item \emph{Gumbel noise.} IID and regime R converge to the same kind of limit, a tilt at the same $\beta=1/b$, but of different base laws: $\pi_{\mathrm{root}}$ for IID and the mutation equilibrium $\pi_0$ for R. Iteration wins asymptotically if and only if $\kappa_0'(1/b)>\kappa_{\mathrm{root}}'(1/b)$, that is, if the tilted mutation equilibrium beats tilted fresh samples.
\item \emph{Gaussian noise and unbounded Gaussian quality under $\pi_{\mathrm{root}}$.} IID grows without bound. If R has a finite stationary mean, IID eventually overtakes it at matched evaluations. When quality is bounded, as in our bin-packing task, this conclusion does not carry over. R can lead only in an intermediate window, and only if the plateau $f^\star(\beta)$ exceeds the IID curve before IID catches up. The late crossover is predicted by solving $\mu+e_{E}s^2/\sqrt{s^2+\sigma^2}=f^\star(\beta)$ for $E$; this gives a \emph{double crossover} (IID $\to$ R $\to$ IID) when SNR is high.
\end{itemize}
In practice the late crossover can lie beyond any realistic budget when the mutation drift is large ($\mu_0\gg\mu$), which is the regime in which LLM-guided evolution is used.

\section{Experiments}
\subsection{Simulation}\label{sec:sim}
We use the autoregressive model of Example~\ref{ex:ar} (details in Appendix~\ref{app:sims}).

\paragraph{The temperature.} Table~\ref{tab:beta} reports the long-run mean of regime R divided by the SNR $d$ under Gaussian noise. As the step size shrinks ($\rho\to1$) it converges to $2/\sqrt\pi=1.128$ for every $d$, as Theorem~\ref{thm:smallstep} predicts, and not to the probit-logit constant $1.2035$. Larger steps raise the plateau, more so at high SNR. Under Gumbel noise the stationary mean equals $1/b$ to within Monte Carlo error for every $\rho$ (Lemma~\ref{lem:barker}).

\begin{table}[h]\centering
\caption{Mean quality of regime R divided by SNR $d$, Gaussian noise: average over the second half of runs of $60/(1-\rho)+500$ steps. Theorem~\ref{thm:smallstep}: $1.128$ as $\rho\to1$. $^\dagger$Grid stationary mean by power iteration; this chain mixes slowly, and a run of $620$ steps gives only $1.23$ (Section~3.4).}\label{tab:beta}
\begin{tabular}{lcccc}\toprule
$d$ & $\rho=0.5$ & $0.9$ & $0.99$ & $0.998$\\\midrule
0.5 & 1.148 & 1.135 & 1.124 & 1.123\\
1 & 1.188 & 1.142 & 1.127 & 1.127\\
3 & 1.332$^\dagger$ & 1.215 & 1.143 & 1.132\\\bottomrule
\end{tabular}\end{table}

\paragraph{The plateau, including drift.} The plateau computed from the stationary law alone (stationary law on a grid by power iteration, no search) matches simulated search: $1.143$ vs.\ $1.14$ ($d=1,\rho=0.9$), $3.653$ vs.\ $3.66$ ($d=3,\rho=0.9$), and with drift $\delta=0.3$, $4.143$ vs.\ $4.14$ ($d=1,\rho=0.9$). The additive form $\beta+\delta/(1-\rho)$ holds within 2\% across $d\in\{0.5,1\}$, $\rho\in\{0.5,0.9\}$, $\delta\in\{0,0.1,0.3\}$.

\begin{figure}[t]\centering\includegraphics[width=\linewidth]{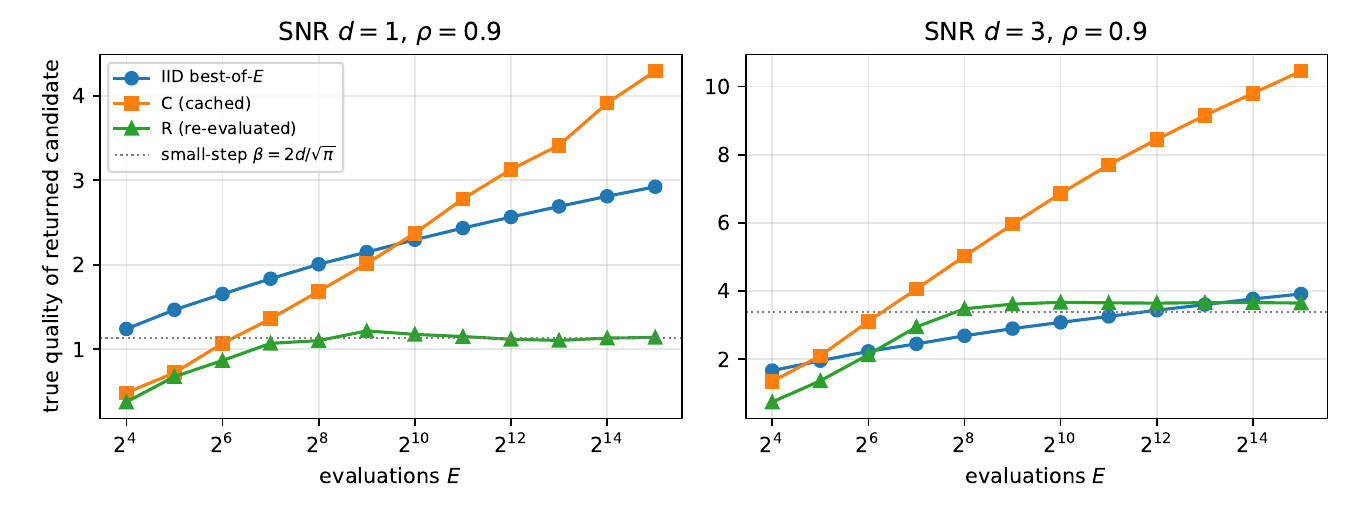}
\caption{True quality returned at matched evaluations (autoregressive model, $\rho=0.9$, Gaussian noise). Re-evaluated search (R) flattens at its plateau; at $d=1$ the dotted small-step value $2d/\sqrt\pi$ is close, at $d=3$ the finite-step plateau ($3.65$) is above it. Cached search (C) keeps rising. IID grows like $\sqrt{2\log E}$.}\label{fig:matched}\end{figure}
\paragraph{Matched evaluations.} Figure~\ref{fig:matched} and Table~\ref{tab:matched} compares the three procedures at $E$ evaluations ($\rho=0.9$, no drift). Regime R is capped at its plateau. At $d=1$ IID is ahead at every budget; at $d=3$ R leads at intermediate budgets and IID overtakes it again between $E=8{,}192$ and $32{,}768$, the double crossover of Section~4. Cached search keeps climbing and leads at large budgets.

\begin{table}[h]\centering
\caption{True quality of the returned candidate at matched evaluations $E$ ($\rho=0.9$, $\delta=0$, 2{,}000 runs).}\label{tab:matched}
\begin{tabular}{rrccc}\toprule
$d$ & $E$ & IID & C (cached) & R (re-evaluated)\\\midrule
1 & 32 & 1.45 & 0.74 & 0.61\\
1 & 512 & 2.16 & 2.03 & 1.17\\
1 & 8192 & 2.69 & 3.52 & 1.14\\
1 & 32768 & 2.92 & 4.26 & 1.13\\
3 & 32 & 1.96 & 2.06 & 1.39\\
3 & 512 & 2.88 & 6.01 & 3.63\\
3 & 8192 & 3.60 & 9.16 & 3.70\\
3 & 32768 & 3.92 & 10.41 & 3.64\\\bottomrule
\end{tabular}\end{table}

\paragraph{Gumbel noise.} With Gumbel noise of scale $b$ and no drift, IID and R converge to the same value $1/b$ (Proposition~\ref{prop:bon}a and Lemma~\ref{lem:barker}): at $E=16{,}384$, $1.00$ vs.\ $1.01$ for $b=1$ and $1.96$ vs.\ $2.01$ for $b=0.5$. With drift $\delta=0.05$ ($\rho=0.9$), R converges to the predicted $1/b+\delta/(1-\rho)$ ($1.51$ vs.\ $1.50$; $2.53$ vs.\ $2.50$) while IID stays at $1/b$, so iteration wins exactly when the mutation equilibrium beats fresh samples.

\subsection{LLM-guided program search: pre-registered tests}\label{sec:llm}
We test the tilt as a prediction for real LLM search loops. In each setting the prediction is computed from calibration data alone and committed to version control before the search is launched; the search then runs on independent seeds. Five registered tests follow: the level of the prediction (test~1), a relation between its error and inheritance found after the first five settings (test~2), slow mixing as an account of the error (test~3), the role of the start program (test~4), and a second domain, prompt optimisation, in which the bin-packing relation was tested out of sample (test~5, Section~\ref{sec:prompt}). Table~\ref{tab:hyp} lists every registered hypothesis, its rule and its verdict, together with the robustness checks of this section. All levels we report are levels over a fixed window of a finite run; we call them plateaus only where runs from different starts agree.

\paragraph{Tasks.} \emph{Bin packing (BP).} A program is a Python priority function for online bin packing (capacity 100; 300 items per instance drawn as $45\cdot\mathrm{Weibull}(3)$). Its score on an instance is $L_1/\text{bins}$, where $L_1$ is the volume lower bound, and its true quality $f$ is the mean over a fixed set of 200 instances. Search starts from first fit ($f=0.9430$). \emph{Travelling salesman (TSP).} A program is a priority function that picks the next city of a tour on 60 uniform random cities. Its score on an instance is the nearest-neighbour tour length divided by its own tour length, so nearest neighbour, the starting program, has $f=1$; $f$ is the mean over 200 fixed instances. TSP scores are real-valued, without the integer bin counts that make BP scores discrete, but ties still occur: on calibration pairs $3.3\%$ of per-instance differences are exactly zero and $0.3\%$ of pairs have identical score vectors (\texttt{ties\_diag.py}). True quality is measured on the same 200 instances that minibatches are drawn from: we test the search dynamics on a finite task pool, not generalisation to new instances.

\paragraph{Proposers.} A mutation sends the current program to an LLM with a fixed instruction and takes the returned program as the child; every proposal is a fresh sample. Seven settings vary the proposer and the instruction: on BP, Claude Haiku~5.5 and Claude Sonnet~5.5, each with a neutral instruction (``write an improved or alternative version''), a small-edit instruction and a rewrite instruction; on TSP, Haiku with the neutral instruction. The Sonnet settings with the small-edit and rewrite instructions were added for test~2. Prompts, sampling and failure handling are in Appendix~\ref{app:llm}. A child that fails to run is never accepted.

\paragraph{Protocol.} For each setting:
(1) \emph{Calibration.} 24 no-selection walks of 100 steps, in which every proposal is accepted, and 600 fresh children of the starting program. The second halves of the walks (1{,}224 programs) estimate the mutation equilibrium $\pi_0$. The difference scale $\tau$ is the median, over walk parent--child pairs, of the standard deviation of per-instance score differences.
(2) \emph{Prediction.} $\beta(m)=1.596\sqrt m/\tau\cdot\sqrt{(N-1)/(N-m)}$, where the last factor corrects for minibatches drawn without replacement from $N=200$ instances, and the tilt prediction $\E_{\pi_0e^{\beta f}}f$ is estimated by reweighting the walk sample. We also record the effective sample size of the weights, $\mathrm{ESS}=(\sum w_i)^2/\sum w_i^2$, and a bootstrap error over walks. The prediction is committed before the search starts.
(3) \emph{Search.} Regime R with $m\in\{1,4,16,64\}$, 6 seeds each, 150 steps (300 for four settings in test~3). The observed level is the mean true quality of the current incumbent over steps 100--150, with the standard error over seeds. A step costs two minibatch evaluations, i.e.\ $2m$ instance evaluations, so comparisons across $m$ at equal steps are not at equal cost; we compare levels across $m$ and costs only within each $m$.
(4) \emph{From above.} For BP with the neutral and small-edit instructions (Haiku and Sonnet), regime R is also started from one program of quality $0.9559$, above every prediction, which the exploratory search of Section~\ref{sec:llm-explore} had selected ($m\in\{1,4,16\}$, 6 seeds, 150 steps; 300 for the small-edit instruction).
(5) \emph{Independent sampling.} Best of $E$ fresh children of the starting program, selected by a size-$m$ minibatch, with $E=300$ equal to the evaluations used by R; the $E$ children are drawn without replacement from the 600 calibration children (failed children, which are never accepted, excluded), so the estimate is unbiased for $E$ fresh valid samples given that pool. The pool itself is a sample, and the reported error does not include its variability.

\begin{table}[!htbp]\centering\footnotesize
\caption{Test~1. ``Pre-reg.\ pred.'': tilt prediction from calibration only ($\pm$ bootstrap error over walks); ESS: effective sample size of the reweighting. ``Observed'': mean true quality of the incumbent over steps 100--150, $\pm$ s.e.\ over 6 seeds; $z$: (observed $-$ prediction)/combined s.e. ``From above'': same, started at quality $0.9559$. ``Lift'': (observed $-$ walk mean)/(prediction $-$ walk mean), in parentheses where ESS $<30$, since the prediction there is not informative. IID for the small-edit settings is omitted: 599 (Haiku) and 600 (Sonnet) of the 600 fresh children have identical quality.}\label{tab:prereg}
\setlength{\tabcolsep}{2.5pt}
\begin{tabular}{llrcrcrcrc}\toprule
setting & $m$ & $\beta$ & pre-reg.\ pred. & ESS & observed & $z$ & from above & lift & IID at $E$\\\midrule
BP, Haiku, neutral & 1 & 281 & $0.9509\pm0.0001$ & 480 & $0.9500\pm0.0005$ & $-1.7$ & $0.9519\pm0.0005$ & 88\% & 0.9491\\
 & 4 & 566 & $0.9525\pm0.0002$ & 188 & $0.9512\pm0.0007$ & $-1.9$ & $0.9525\pm0.0006$ & 86\% & 0.9496\\
 & 16 & 1168 & $0.9541\pm0.0002$ & 48 & $0.9533\pm0.0010$ & $-0.9$ & $0.9543\pm0.0002$ & 92\% & 0.9512\\
 & 64 & 2716 & $0.9551\pm0.0001$ & 10 & $0.9541\pm0.0006$ & $-1.5$ & -- & (91\%) & 0.9529\\
\addlinespace
BP, Haiku, small edit & 1 & 424 & $0.9512\pm0.0006$ & 397 & $0.9499\pm0.0010$ & $-1.1$ & $0.9503\pm0.0007$ & 76\% & --\\
 & 4 & 855 & $0.9531\pm0.0007$ & 82 & $0.9510\pm0.0005$ & $-2.4$ & $0.9508\pm0.0007$ & 71\% & --\\
 & 16 & 1765 & $0.9548\pm0.0009$ & 13 & $0.9525\pm0.0006$ & $-2.2$ & $0.9509\pm0.0005$ & (75\%) & --\\
 & 64 & 4107 & $0.9554\pm0.0008$ & 5 & $0.9531\pm0.0006$ & $-2.3$ & -- & (76\%) & --\\
\addlinespace
BP, Haiku, rewrite & 1 & 221 & $0.9508\pm0.0001$ & 801 & $0.9513\pm0.0009$ & $+0.6$ & -- & 111\% & 0.9519\\
 & 4 & 445 & $0.9518\pm0.0001$ & 472 & $0.9521\pm0.0002$ & $+1.1$ & -- & 105\% & 0.9532\\
 & 16 & 919 & $0.9532\pm0.0001$ & 142 & $0.9539\pm0.0004$ & $+1.8$ & -- & 111\% & 0.9545\\
 & 64 & 2137 & $0.9545\pm0.0002$ & 19 & $0.9549\pm0.0002$ & $+1.4$ & -- & (105\%) & 0.9550\\
\addlinespace
BP, Sonnet, neutral & 1 & 401 & $0.9523\pm0.0003$ & 534 & $0.9513\pm0.0008$ & $-1.1$ & $0.9531\pm0.0007$ & 83\% & 0.9487\\
 & 4 & 808 & $0.9534\pm0.0003$ & 226 & $0.9516\pm0.0005$ & $-2.9$ & $0.9512\pm0.0007$ & 73\% & 0.9493\\
 & 16 & 1668 & $0.9544\pm0.0003$ & 52 & $0.9532\pm0.0006$ & $-1.7$ & $0.9528\pm0.0009$ & 85\% & 0.9511\\
 & 64 & 3881 & $0.9552\pm0.0003$ & 9 & $0.9541\pm0.0003$ & $-2.8$ & -- & (87\%) & 0.9525\\
\addlinespace
BP, Sonnet, small edit & 1 & 351 & $0.9521\pm0.0005$ & 626 & $0.9509\pm0.0007$ & $-1.5$ & $0.9531\pm0.0004$ & 56\% & --\\
 & 4 & 708 & $0.9535\pm0.0003$ & 290 & $0.9511\pm0.0008$ & $-2.7$ & $0.9519\pm0.0004$ & 42\% & --\\
 & 16 & 1462 & $0.9543\pm0.0003$ & 122 & $0.9525\pm0.0006$ & $-2.8$ & $0.9523\pm0.0008$ & 64\% & --\\
 & 64 & 3401 & $0.9549\pm0.0003$ & 30 & $0.9540\pm0.0002$ & $-2.5$ & -- & 84\% & --\\
\addlinespace
BP, Sonnet, rewrite & 1 & 209 & $0.9486\pm0.0002$ & 697 & $0.9509\pm0.0006$ & $+3.8$ & -- & 155\% & 0.9525\\
 & 4 & 421 & $0.9505\pm0.0002$ & 268 & $0.9500\pm0.0006$ & $-0.7$ & -- & 92\% & 0.9535\\
 & 16 & 868 & $0.9531\pm0.0003$ & 41 & $0.9539\pm0.0004$ & $+1.5$ & -- & 109\% & 0.9545\\
 & 64 & 2019 & $0.9548\pm0.0003$ & 7 & $0.9549\pm0.0002$ & $+0.3$ & -- & (101\%) & 0.9552\\
\addlinespace
TSP, Haiku, neutral & 1 & 24 & $0.9457\pm0.0051$ & 321 & $0.9378\pm0.0127$ & $-0.6$ & -- & 92\% & 1.0101\\
 & 4 & 48 & $0.9866\pm0.0110$ & 31 & $0.9923\pm0.0183$ & $+0.3$ & -- & 104\% & 1.0221\\
 & 16 & 100 & $1.0359\pm0.0170$ & 2 & $1.0115\pm0.0148$ & $-1.1$ & -- & (87\%) & 1.0306\\
 & 64 & 232 & $1.0464\pm0.0144$ & 1 & $1.0680\pm0.0118$ & $+1.2$ & -- & (111\%) & 1.0427\\
\bottomrule\end{tabular}

\end{table}

\begin{figure}[!htbp]\centering\includegraphics[width=\linewidth]{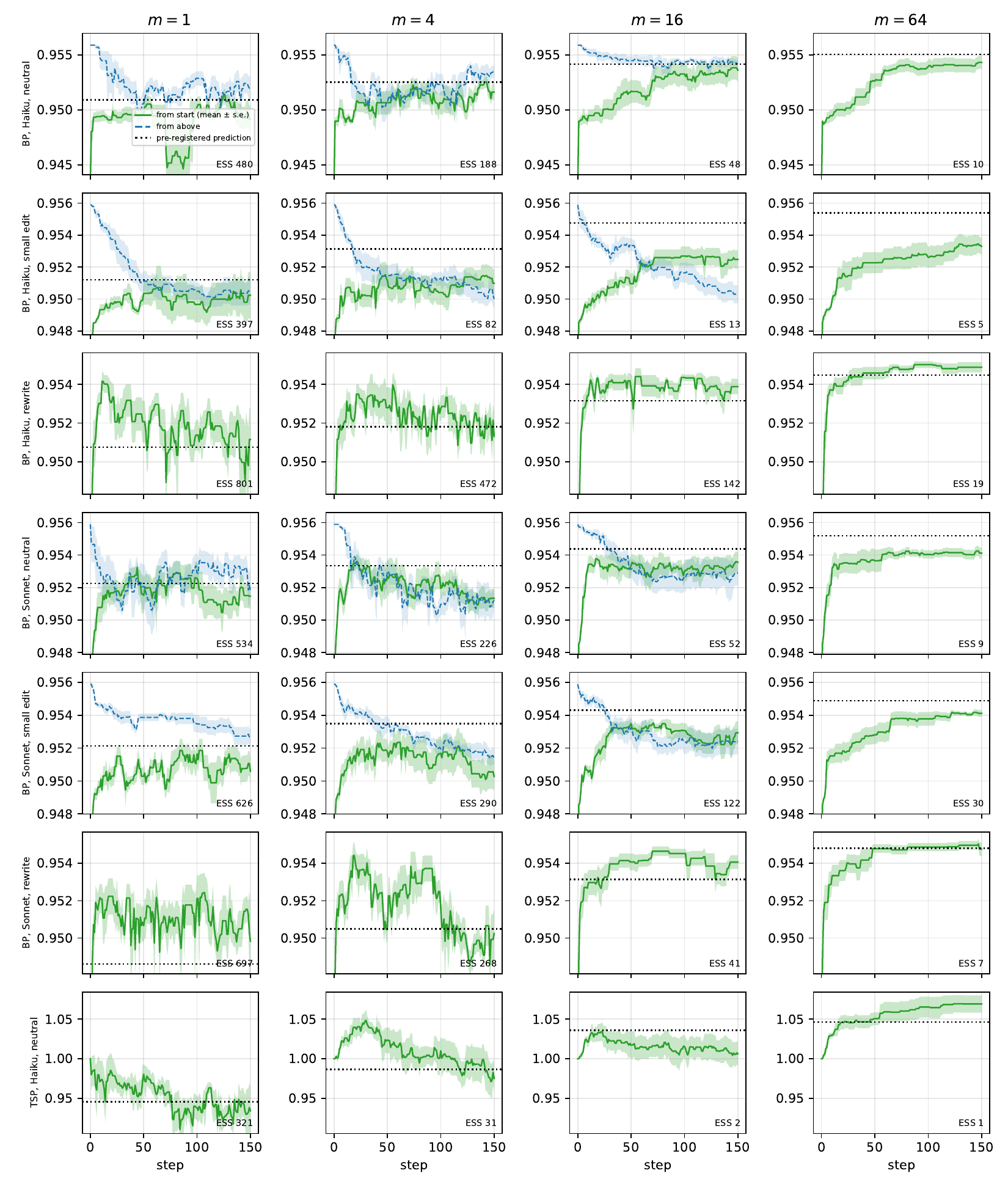}
\caption{True quality of the incumbent in regime R (mean $\pm$ s.e.\ over 6 seeds) for each setting and minibatch size. Dotted: pre-registered prediction. Dashed: runs started from above. ESS of the reweighting in each panel; where it is in single digits the prediction is close to the largest walk sample and does not test the shape of $\pi_0$.}\label{fig:prereg}\end{figure}

\begin{table}[!p]\centering\footnotesize
\caption{Every registered hypothesis, its decision rule and its verdict. Test 1 registered predictions without a numerical pass/fail rule; tests 2--5 registered rules before calibration (tests 2 and 5), before any step beyond 150 (test 3 and H6-300/H10) or before any step of the new runs (test 4). Test 5 is the prompt-optimisation domain (Section~\ref{sec:prompt}); its $L$ rests on the one cell with ESS $\ge30$ ($m=4$) and its se is over 6 seeds. $L$: attained fraction of the predicted lift; $\rho$: parent--child quality correlation of the calibration walks; se: standard error over 6 seeds (test 4: across starts, 6 per group). $D$: mean drop of walk-visited minus search-selected starts. Robustness checks (post-hoc, Section~\ref{sec:llm}) correct two features of the registered analyses: cells that share proposals within a seed (H1q) and pooled proposals that over-weight hard-to-improve search parents (H2). Test 5 scores are frozen single evaluations of each prompt.}\label{tab:hyp}
\begin{tabular}{@{}l>{\raggedright\arraybackslash}p{2.4cm}>{\raggedright\arraybackslash}p{4.3cm}>{\raggedright\arraybackslash}p{4.3cm}>{\raggedright\arraybackslash}p{2.5cm}@{}}\toprule
Test & Hypothesis & Rule & Result & Verdict\\\midrule
1 & plateau level = tilt & none (report $z$) & editing: 16/16 cells below, $z\in[-2.9,-0.9]$; rewrite: 7/8 above & level fails\\
1 & ordering by $m$ & none & six of seven settings & sanity check\\\midrule
2 & H1a (rewrite) & fail if $L+2\,\mathrm{se}<0.95$ & $L=1.19\pm0.06$ & pass\\
2 & H1b (small edit) & fail if $L-2\,\mathrm{se}>0.85$ & $L=0.62\pm0.08$ & pass\\
2 & H1q (both) & $|L-\hat L(\rho)|\le2\sqrt{0.020^2+\mathrm{se}^2}$ & $0.04$ (tol.\ $0.12$); $0.15$ (tol.\ $0.17$). Seed-blocked se: tol.\ $0.15$, $0.10$. Approx.\ propagation: 95\% CI $[-0.11,0.21]$, $[-0.36,0.07]$ & pass, pass. Seed-blocked: pass, \textbf{fail}\\
2 & H2 (both) & search vs.\ walk P(better) & $0.02$ vs.\ $0.04$; $0.17$ vs.\ $0.28$. First proposals only: $0.02$ vs.\ $0.04$; $0.22$ vs.\ $0.26$ & pass, pass. First proposals: pass, \textbf{fail} ($0.046<0.05$)\\\midrule
3 & H3 (small edit, 2 settings) & mixing rejected if $\Delta+2\,\mathrm{se}<\Delta_{\rm null}/2$; persistence if $\Delta-2\,\mathrm{se}>\Delta_{\rm null}/2$ & $\Delta=-0.06\pm0.12$; $+0.11\pm0.10$ & inconclusive (2/2)\\
3 & H4 (neutral, 2 settings) & mixing rejected if $L(W_{300})+2\,\mathrm{se}<0.95$ & Haiku $0.93\pm0.03$; Sonnet $0.82\pm0.06$ & null model not rejected; rejected (boundary)\\
3 & H5 (from above, 4 cells) & mixing rejected if level$+2\,\mathrm{se}<$ pred.$-0.0005$ & all 4 cells $0.0024$--$0.0043$ below pred. & null model rejected (4/4; 3 already at 150 steps)\\\midrule
4 & H6 (start, steps 1--50) & selection effect if $D-2\,\mathrm{se}>0$; fail if $D+2\,\mathrm{se}<0$ & $D=0.0003\pm0.0004$ & inconclusive\\
4 & H6-late (100--150, sec.) & same rule & $D=0.0016\pm0.0004$ & pass\\
4 & H6-300 (250--300) & apart if $D-2\,\mathrm{se}>0$; converged if $|D|+2\,\mathrm{se}<0.0005$ & $D=0.0011\pm0.0005$ & apart (narrowly)\\
4 & H10 (250--300) & above / at / below pred.\ ($0.9535$), $\pm2\,\mathrm{se}$ incl.\ pred.\ error & walk $0.9535\pm0.0004$; search $0.9525\pm0.0004$ & walk at (compatible, not equivalent); search below (narrowly)\\\midrule
5 & H7 (ordering) & $m=64$ above $m=4$ & $0.963$ vs.\ $0.756$; $0.903$ vs.\ $0.781$ & holds (sanity check)\\
5 & H8 (direction) & $L$(bold) $>$ $L$(small): pass if diff $-2\,\mathrm{se}>0$, fail if diff $+2\,\mathrm{se}<0$ & diff $=-4.20\pm1.38$; bootstrap $[-18.6,-0.9]$ & \textbf{fail}\\
5 & H9 (transfer, 2 settings) & $|L-\hat L(\rho)|\le2\sqrt{0.065^2+\mathrm{se}^2}$ & small: $5.98$ vs.\ $0.80$ (tol.\ $2.64$); bold: $1.78$ vs.\ $1.02$ (tol.\ $0.85$). Calibration bootstrap, $L-\hat L$: $[2.2,19.7]$; $[-0.05,1.90]$ & \textbf{fail}, pass (not rejected)\\\bottomrule
\end{tabular}\end{table}

\paragraph{Test 1: the level.} Table~\ref{tab:prereg} and Figure~\ref{fig:prereg}.
(i) \emph{Ordering.} The observed level rises with $m$ in six of seven settings; in Sonnet's rewrite setting $m=1$ lies above $m=4$ ($0.9509$ against $0.9500$), and in two Sonnet settings $m=1$ and $m=4$ are equal within error. Since any model in which a less noisy gate selects better predicts this, we treat it as a sanity check, not as support for the tilt. On BP the rise from $m=1$ to $m=64$ is $0.0028$--$0.0041$; on TSP, where $\pi_0$ is ten times wider, it is $0.13$.
(ii) \emph{Level on bin packing.} With the neutral and small-edit instructions (both models), the pre-registered prediction is too high: all 16 cells lie below it, by $0.0008$ to $0.0024$ ($z$ between $-0.9$ and $-2.9$). Measured as the fraction of the predicted lift over the walk mean, R attains $86$--$92\%$ (Haiku, neutral), $71$--$76\%$ (Haiku, small edit), $73$--$87\%$ (Sonnet, neutral) and $42$--$84\%$ (Sonnet, small edit). With the rewrite instruction the prediction is too low rather than too high: all four Haiku cells lie above it ($z$ between $+0.6$ and $+1.8$; $105$--$111\%$ of the lift), and so do three of the four Sonnet cells ($155\%$, $109\%$ and $101\%$ at $m=1,16,64$, the last with ESS $7$; $92\%$ at $m=4$).
(iii) \emph{Runs from above.} With Haiku and the neutral instruction they settle at $0.9519$, $0.9525$ and $0.9543$ for $m=1,4,16$, so the runs from below and from above bracket the prediction ($0.9509$, $0.9525$, $0.9541$): part of the shortfall there is incomplete mixing in 150 steps. With Sonnet and the neutral instruction both lie below the prediction at $m=4$ ($0.9516$ from below, $0.9512$ from above, prediction $0.9534$) and at $m=16$ ($0.9532$, $0.9528$, prediction $0.9544$). With the small-edit instruction the runs from above fall below the runs from below or to their level: with Haiku to $0.9503$, $0.9508$ and $0.9509$ ($m=1,4,16$; predictions $0.9512$, $0.9531$, $0.9548$), with Sonnet to $0.9519$ and $0.9523$ at $m=4$ and $16$ (predictions $0.9535$ and $0.9543$); at Sonnet $m=1$ the run from above is still falling.
(iv) \emph{TSP.} The starting program lies far above the mutation equilibrium (walk mean $0.852$), so these chains approach the prediction from above. The observed levels ($0.938$, $0.992$, $1.012$, $1.068$) are within $1.2$ combined standard errors of the predictions, but the seed errors are large ($0.012$--$0.018$), and at $m=16$ and $64$ the ESS is $2$ and $1$, so the test is uninformative there.
(v) \emph{Iterating versus independent sampling.} On BP with the neutral instruction (Haiku or Sonnet), R is above best-of-$E$ fresh sampling at every $m$, by $0.0009$--$0.0026$. With the rewrite instruction best-of-$E$ is ahead at every $m$, by $0.0001$--$0.0011$ (Haiku) and $0.0003$--$0.0035$ (Sonnet; the largest gap is at $m=4$, where R's late window dips). On TSP the order reverses for $m\le16$: best-of-$E$ reaches $1.010$--$1.031$ while R sits at $0.94$--$1.01$, because mutation degrades programs on average (walk mean $0.852$ against a start of $1$) while fresh children of nearest neighbour have a heavy upper tail. The pattern resembles the Gumbel case of Section~4, where iteration wins exactly when the tilted mutation equilibrium beats the tilted root distribution; with near-Gaussian, pair-dependent noise and finite runs, that statement is a heuristic reading here, not a consequence.

\paragraph{How sharp is test 1?} Without selection ($\beta=0$) the prediction is the walk mean, $0.944$--$0.947$ on BP; a chain that reached the best walk sample would be at $0.9554$--$0.9556$. The observed levels lie between these; with the editing instructions they fall short of the tilt by $8$--$58\%$ of the lift, and with the rewrite instruction they exceed it in seven of eight cells. Doubling $\beta$ raises the BP prediction by $0.0006$--$0.0019$, so the test can resolve a factor-two error in $\beta$. Two accounts of the shortfall are consistent with the levels alone: a correct temperature applied to a base law worse than $\pi_0$ near the top, or a lower effective temperature. A check of the acceptance rule on walk pairs, simulating the paired gate with minibatches drawn without replacement, finds the exact acceptance log-odds at or above the slope of the Gaussian approximation, within bootstrap error, at every $m$ and in every BP setting (slope ratios $0.98$--$2.11$; Appendix~\ref{app:diag}). This removes a flatter acceptance function as an explanation, on the pairs a walk visits; it does not address heterogeneity of $\tau_{xy}$ over the pairs a search visits.

\paragraph{Test 2: inheritance.} After the first five settings, post-hoc checks (Appendix~\ref{app:diag}) found that neither these re-estimates of $\tau$ nor a projected kernel that drops reversibility removed the error, and showed that the attained fraction of the lift ordered with the parent--child quality correlation of the calibration walks: $0.13$ (Haiku rewrite), $0.61$ (Haiku neutral), $0.84$ (Sonnet neutral), $0.91$ (Haiku small edit), against $105$--$111\%$, $86$--$92\%$, $73$--$87\%$ and $71$--$76\%$. Before calibrating two new settings, Sonnet with the rewrite and with the small-edit instruction, we committed four hypotheses, their decision rules and the evaluation script. Let $L$ be the attained fraction of the predicted lift, averaged over the cells with ESS at least $30$, and $\rho$ the parent--child quality correlation on calibration walk pairs. On the four BP settings, $L=1.148-0.434\,\rho$ with residual s.d.\ $0.020$. \emph{H1a}: with the rewrite instruction, $L\ge0.95$ (fail if $L+2\,\mathrm{se}<0.95$; void if $\rho>0.3$). \emph{H1b}: with the small-edit instruction, $L\le0.85$ (fail if $L-2\,\mathrm{se}>0.85$; void if $\rho<0.7$). \emph{H1q}: $|L-\hat L(\rho)|\le2\sqrt{0.020^2+\mathrm{se}^2}$, where $\hat L(\rho)$ is computed from the calibration $\rho$ and committed with the tilt prediction, before the search. \emph{H2}: for parents of quality in $[0.952,0.957)$, the probability that the child is better is at least $0.05$ lower for search parents than for walk parents if $\rho>0.7$, and within $0.05$ if $\rho<0.3$.

Calibration gave $\rho=-0.01$ (rewrite) and $0.88$ (small edit), hence $\hat L=1.15$ and $0.77$. Search gave $L=1.19$ and $0.62$. Under the registered rules all four hypotheses pass: H1a and H1b by wide margins, H1q with differences of $0.04$ (tolerance $0.12$) and $0.15$ (tolerance $0.17$), and H2 with $0.17$ vs.\ $0.28$ for the small-edit setting and $0.02$ vs.\ $0.04$ for the rewrite setting. Two of these passes do not survive a correction of the registered analysis. \emph{H1q.} The registered standard error of $L$ treats the cells $m=1,\dots,64$ as independent, but within a seed they share proposals. Averaging the lift over cells within each seed and taking the standard error over seeds gives $\mathrm{se}=0.070$ and $0.045$ instead of $0.058$ and $0.084$; with the registered rule the tolerance becomes $0.15$ and $0.10$, and the small-edit setting fails ($0.15>0.10$). Propagating, approximately, also the uncertainty of the two settings' calibration (block bootstrap over walks, through $\pi_0$, $\tau$ and $\rho$) and of the four-point fit (normal draws from its fixed-design least-squares covariance, which has two residual degrees of freedom; the four training settings are not resampled) gives 95\% intervals for $L-\hat L$ of $[-0.11,+0.21]$ (rewrite) and $[-0.36,+0.07]$ (small edit), dominated by the calibration walks in the small-edit setting. The quantitative agreement is therefore neither established nor refuted; the directional hypotheses H1a and H1b are unaffected. \emph{H2.} The registered statistic pools all proposals from a parent. A search incumbent receives proposals until one is accepted, so parents that are hard to improve receive more proposals and are over-represented; walk parents receive one proposal per visit. In a toy model in which search and walk parents have the same distribution of improvement probabilities, the pooled rate is $0.21$ for search parents against $0.25$ for walk parents (\texttt{toy\_sampling\_bias.py}). Using only the first proposal from each parent in each run, which is drawn before any stopping decision, and excluding the chains from above (which the registered statistic included), the small-edit rates are $0.22$ vs.\ $0.26$ ($n=227$ and $413$; a difference of $0.046$, short of the registered $0.05$) and the rewrite rates $0.02$ vs.\ $0.04$. Table~\ref{tab:hyp} reports both versions.
\emph{Post-hoc six-setting fit.} Refitting on all six BP settings, after the test, gives $L=1.18\,(\pm0.05)-0.52\,(\pm0.07)\,\rho$, with residual s.d.\ $0.065$ and correlation $-0.96$; the line crosses $L=1$ at $\rho\approx0.34$. With six settings, two models and one task, we read it as an empirical regularity about finite-budget runs: the less a child inherits, the closer the tilt comes to the observed level after 150 steps and, for children nearly independent of the parent, the more it underestimates it.

\paragraph{What the regularity does and does not say.} Under the Barker acceptance condition of Lemma~\ref{lem:barker}, the tilt of $\pi_0$ is the stationary law of every proposal kernel that is reversible with respect to $\pi_0$ on program space, whatever the program's internal structure, so, if the chain is also irreducible and the tilt is therefore its unique stationary law, inheritance alone cannot move a stationary level away from the tilt; it can lengthen the time to reach it. The actual minibatch gates meet that condition only approximately (Section~3.2), so a deviation can also come from finite-step, non-logistic or pair-dependent acceptance; and our predictors read the kernel through scalar quality only, from unselected programs. Three post-hoc analyses (Appendix~\ref{app:diag}) bear on which of these applies. (a) With the rewrite instruction, children depend on the parent's lineage depth, and the calibration walks are deeper than the search incumbents, so $\pi_0$ estimated from the walks is worse than the proposal law the search samples; tilting early walk samples instead removes the underestimate, but that correction was found on the same data. (b) The evidence that search-selected parents are harder to improve than walk parents of the same quality is weaker than the pooled rates suggested: with first proposals only, the odds ratio of improvement (search vs.\ walk, adjusted for parent quality, parents in $[0.952,0.957)$) is $1.05$ (Haiku neutral), $0.51$ (Haiku small edit, 95\% interval $[0.28,0.92]$), $0.74$ (Sonnet neutral, $[0.54,1.01]$) and $0.80$ (Sonnet small edit, $[0.54,1.19]$), with intervals clustered by parent. First proposals of different parents along one run are not independent, and with errors clustered by run instead the Haiku small-edit interval is $[0.16,1.59]$ (6 search and 13 walk runs) and no setting is significant at 5\%; with this few clusters the normal approximation is itself rough. We read (b) as exploratory. (c) A reversible null model with the walks' autocorrelation reproduces the size of the shortfall for small edits after 150 steps but not its time course, not the shortfall for neutral edits, and not the runs from above, which it keeps at the prediction. Tests~3 and~4 were registered to look at the dynamics directly.

\paragraph{Test 3: slow mixing (300 steps).} Under the null model of (c) more steps raise the attained fraction and keep runs from above near the prediction. After (c), and before any step beyond 150 was generated, we committed a third test with its decision rules and evaluation script: the runs of four settings (Haiku and Sonnet, neutral and small-edit instructions) and the small-edit runs from above are continued to 300 steps with the same seeds, and steps 250--300 ($W_{300}$) are compared with steps 100--150 ($W_{150}$). \emph{H3} (small edit): with $\Delta=L(W_{300})-L(W_{150})$ per seed, slow mixing is rejected if $\Delta+2\,\mathrm{se}<\Delta_{\rm null}/2$ and persistence if $\Delta-2\,\mathrm{se}>\Delta_{\rm null}/2$, where $\Delta_{\rm null}=0.21$ (Haiku) and $0.19$ (Sonnet) is the null model's rise. \emph{H4} (neutral; null model $L(W_{300})=1.00$): slow mixing is rejected if $L(W_{300})+2\,\mathrm{se}<0.95$. \emph{H5} (from above, $m=4,16$): slow mixing is rejected if the level over $W_{300}$ plus $2\,\mathrm{se}$ is below the prediction minus $0.0005$; the null model's runs from above are within $0.0002$ of the predictions there. A power calculation, committed after launch and before any result, found that H3 was unlikely to decide with six seeds and that H5 carried the test.

\begin{figure}[!htbp]\centering\includegraphics[width=\linewidth]{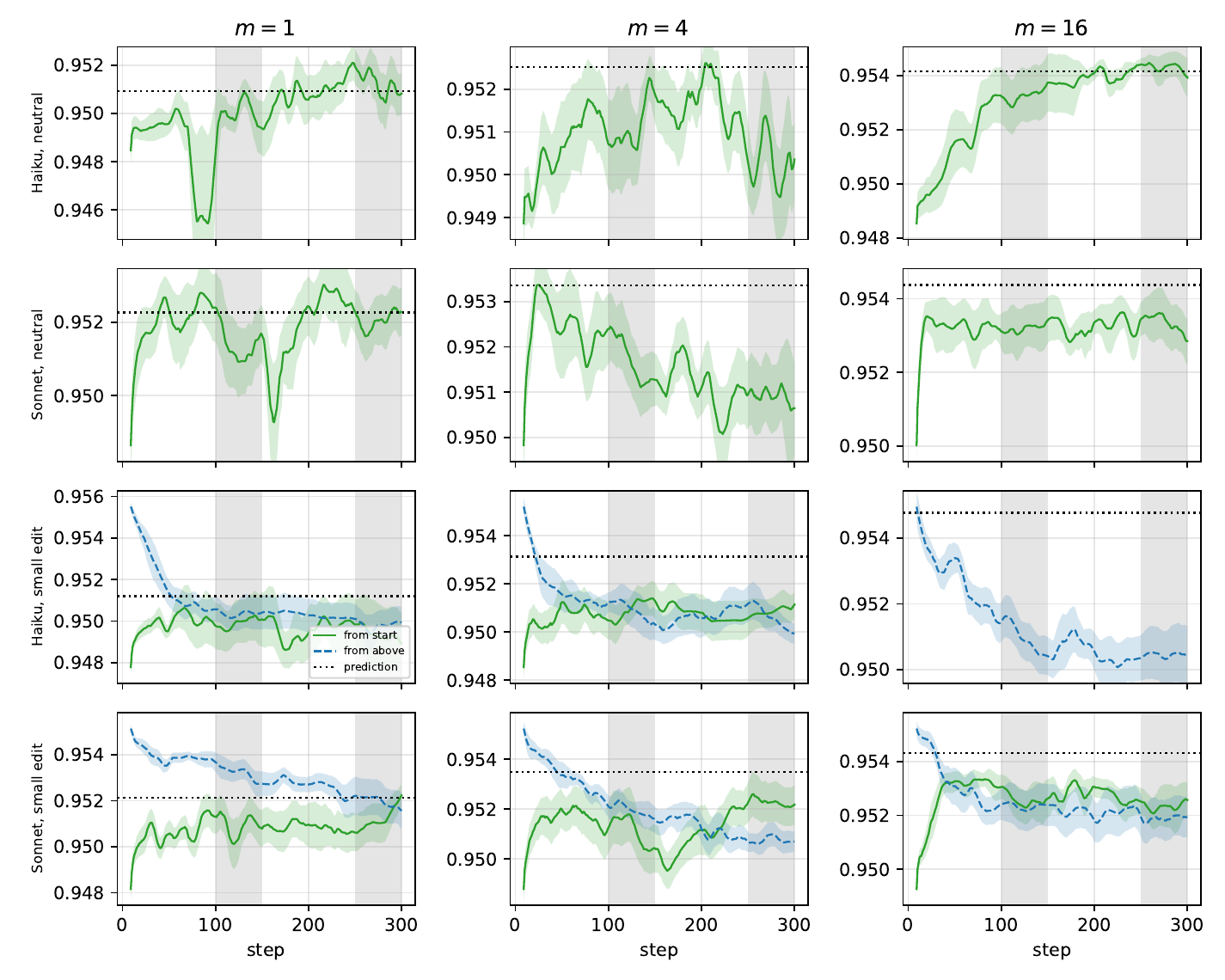}
\caption{Test~3: incumbent true quality over 300 steps (mean $\pm$ s.e. over 6 seeds, 10-step moving average). Solid: from the start program; dashed: from the single high-quality start program (small-edit settings only); dotted: pre-registered prediction. Shaded: windows $W_{150}$ and $W_{300}$. Haiku with the small-edit instruction has no $m=16$ run from below continued to 300 steps.}\label{fig:t300}\end{figure}

In every continued run the first 151 steps reproduce the stored run (Figure~\ref{fig:t300}). \emph{H5 rejects the null model in all four cells.} Over $W_{300}$ the runs from above are at $0.9505\pm0.0005$ and $0.9505\pm0.0008$ (Haiku, $m=4,16$; predictions $0.9531$, $0.9548$) and $0.9508\pm0.0004$ and $0.9519\pm0.0007$ (Sonnet; $0.9535$, $0.9543$), each lower than over $W_{150}$ ($0.9508$, $0.9509$, $0.9519$, $0.9523$). This verdict was largely settled before the continuation: the $W_{150}$ levels were already below the H5 threshold in three cells and at it in the fourth. What the continuation adds is that these runs do not return within another 150 steps, and that runs from above and from below meet or cross below the prediction: for Haiku at $m=4$ they meet ($0.9505$ and $0.9509$); for Sonnet at $m=16$ they are within $0.0004$ of each other, $0.0024$ below the prediction; for Sonnet at $m=4$ the run from above is below the mean of the runs from below in five of six seeds. Not every cell behaves this way: for Sonnet at $m=1$ the two runs meet at the prediction ($0.9519$ and $0.9513$ against $0.9521$), and one of the six Haiku $m=16$ runs from above ends at $0.9542$. \emph{H4} rejects the null model for Sonnet only at the boundary ($L(W_{300})=0.82\pm0.06$) and not for Haiku, whose attained fraction rose from $0.88$ to $0.93$. \emph{H3} is inconclusive in both small-edit settings, as the power calculation expected ($\Delta=-0.06\pm0.12$ and $+0.11\pm0.10$); the attained fractions over $W_{300}$ are $0.67\pm0.12$ and $0.73\pm0.07$, against $0.93$ and $0.98$ for the null model. In three of the four settings the runs from below moved toward the prediction between the two windows. H5 rejects one reversible model, unimodal in a latent inherited component, started from one program. It does not reject reversibility on program space. Under the Barker condition a proposal kernel reversible with respect to $\pi_0$ has the tilt as an exact invariant law. If the chain is moreover irreducible, a level below the tilt reached within 300 steps is either not yet stationary (for instance, metastable) or reflects a $\hat\pi_0$ that misestimates the kernel's equilibrium near the top; a reducible chain can also be stationary within a class of programs it cannot leave; for the actual minibatch gates it may also reflect finite-step, non-logistic or pair-dependent acceptance (Section~3.2).

\paragraph{Test 4: the start program.} Every run from above starts from one program that a search had selected, so tests~1 and~3 cannot tell whether it loses quality because it was selected or because of its quality. After test~3, and before any step of the new runs was generated, we committed a fourth test. For Sonnet with the small-edit instruction we took twelve start programs of matched true quality (mean $0.9544$, range $0.9541$--$0.9547$; prediction at $m=4$: $0.9535$): six visited by unselected calibration walks (one per walk) and six that had been incumbents of the root searches (one per seed). True quality is the score on all instances, so choosing a program inside the band involves no winner's curse in either group. The selected starts had been incumbents of $m=64$ chains (five) and an $m=16$ chain (one), i.e.\ selected by gates with $2$--$5$ times the inverse temperature of the test gate; the registered contrast is selection as such, not selection at $m=4$ (origins in Appendix~\ref{app:llm}). Each start was run in regime R at $m=4$ for 150 steps with seeds 60 and 61. Within a seed all twelve starts draw the same sequence of minibatch indices until their chains desynchronise, which helps the paired contrast but means the 24 runs are not independent; the unit of analysis is the start. For each start, the drop over a window is the mean true quality of the incumbent minus the start's quality, averaged over the two seeds; $D$ is the mean drop of the walk-visited starts minus that of the search-selected starts, with the standard error across starts. \emph{H6} (primary, steps 1--50): selection predicts $D>0$; pass if $D-2\,\mathrm{se}>0$, fail if $D+2\,\mathrm{se}<0$, inconclusive otherwise. The same rule over steps 100--150 was registered as secondary, and a regression of the drop on start type and lineage depth without a rule. The walk-visited starts are deeper in their lineage (mean $43$ generations against $16$), which by (a) would work against the hypothesis if the depth effect of the rewrite setting carried over. The same design with Haiku and the small-edit instruction, where only two walks reach the band, was registered as descriptive.

\begin{figure}[!htbp]\centering\includegraphics[width=\linewidth]{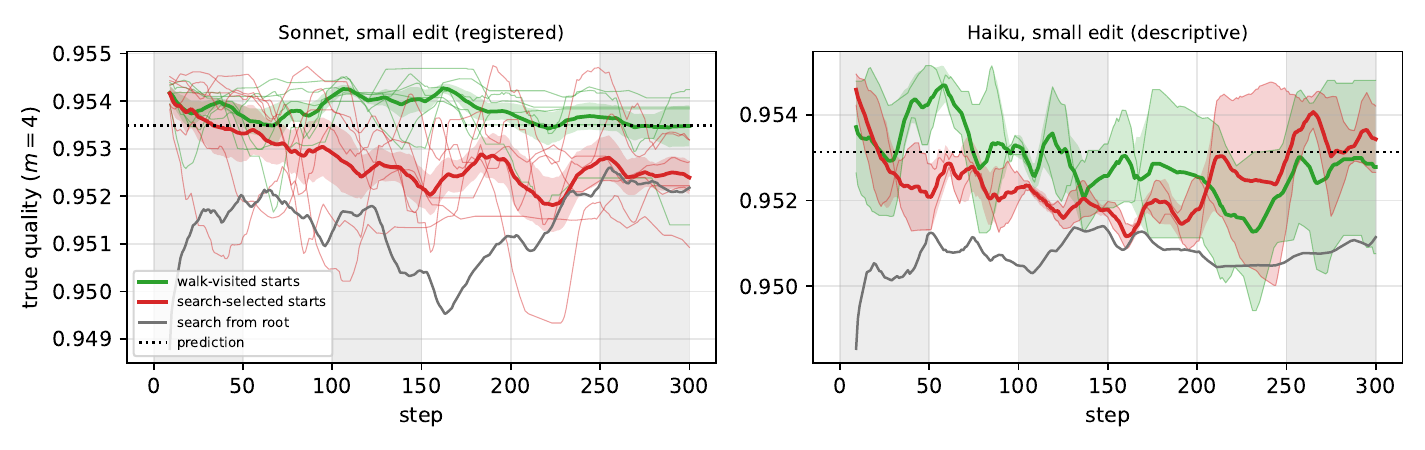}
\caption{Test~4: incumbent true quality in regime R at $m=4$ from start programs of matched quality (10-step moving average). Thin lines: single starts (mean of two seeds); thick: group mean $\pm$ s.e.\ over starts (Sonnet: 6 per group; Haiku: 2). Grey: runs from the root (mean of 6 seeds). Dotted: pre-registered prediction. Shaded: the primary (steps 1--50), secondary (100--150) and continuation (250--300) windows. Steps beyond 150 are the registered continuation, rerun after the evaluation fault of Appendix~\ref{app:llm}.}\label{fig:multistart}\end{figure}

\emph{H6 is inconclusive on its primary window.} Over steps 1--50 the walk-visited starts lose $0.0005$ and the search-selected starts $0.0008$ ($D=0.0003\pm0.0004$). \emph{The secondary window passes.} Over steps 100--150 they lose $0.0003$ and $0.0019$ ($D=0.0016\pm0.0004$); in 35 of the 36 cross-group pairs of starts the search-selected start loses more, and the regression gives $0.0019\pm0.0006$ for start type and $-0.000008\pm0.000014$ per generation of depth (Figure~\ref{fig:multistart}). On the 10-step moving average of the drops the gap between the groups exceeds two standard errors intermittently from step 76 and at every step from step 102. Over steps 100--150 the walk-visited starts are at $0.9541$ ($\pm0.0001$ across starts), $0.0006$ above the prediction (about $1.8$ standard errors when the prediction's bootstrap error is included), and the search-selected starts at $0.9525$, near the level of the runs from the root over steps 250--300 ($0.9523$). In the descriptive Haiku arm both groups lose quality: the walk-visited starts fall from $0.9549$ to $0.9528$ and the search-selected starts from $0.9553$ to $0.9519$ over steps 100--150 (prediction $0.9531$; drops $-0.0023$, $-0.0020$ and $-0.0036$, $-0.0031$ for the four starts), so only the sign of the difference matches the Sonnet result; the Haiku starts sit further above their prediction ($0.0017$--$0.0022$) than the Sonnet starts ($0.0009$).
In the registered setting, then, among the tested starts the subsequent loss was associated with origin even after matching their initial measured quality: programs of the same quality that a walk visited kept it for 150 steps. The single-step analysis (b) shows a weaker and less consistent selection effect, so the quantity that separates the groups over a run need not be the probability that the next child is better. The test does not settle the stationary level. Chains of one kernel at one $m$ stand at different levels after 150 steps depending on the start, so the chains have not been shown to mix to a common invariant law within the budget (either at least one group is not yet stationary, or the chain is reducible on the region it visits), and every incumbent of a run from the root is itself a selected program, so the common level below the tilt in test~3 was reached only by chains of selected programs.

\emph{Continuation to 300 steps.} Before any step beyond 150 was generated, we registered two further hypotheses for the same chains, with the readings of each outcome fixed in advance. \emph{H6-300}: over steps 250--300, the groups are ``apart'' if $D-2\,\mathrm{se}>0$, ``reversed'' if $D+2\,\mathrm{se}<0$, ``converged'' if $|D|+2\,\mathrm{se}<0.0005$, and inconclusive otherwise. \emph{H10}: each group's level over steps 250--300 is ``above'', ``at'' or ``below'' the prediction ($0.9535$) by a two-standard-error rule, with the prediction's bootstrap error ($0.0003$) added in quadrature. Every continued run reproduces its first 151 steps. (The numbers below are from a rerun: in the first run of the continuation the evaluation harness failed on the machine it ran on, so no new child was scored and the chains stayed at their incumbents; we repaired the environment, re-scored the stored children and reran with the same command, seeds and rules; Appendix~\ref{app:llm}.) Over steps 250--300 the walk-visited starts are at $0.9535$ (drops of $0.0003$--$0.0026$ from their start) and the search-selected starts at $0.9525$ (drops of $0.0009$--$0.0030$): $D=0.0011\pm0.0005$, so the groups are \emph{apart}, narrowly ($D-2\,\mathrm{se}=0.0001$). The walk-visited group is \emph{at} the prediction in the registered sense ($0.9535\pm0.0004$: within two standard errors of it, which is statistical compatibility, not equivalence; the $\pm2$\,se interval of its difference from the prediction is $[-0.0008,+0.0009]$) and the search-selected group \emph{below} it ($0.9525\pm0.0004$, an upper bound $0.0001$ under the prediction), near the runs from the root over the same window ($0.9523$). The registered reading of this combination is that the start's history still determines the level after 300 steps. We state it more narrowly: among these twelve starts, whose origin was not randomised, origin and level after 300 steps remain associated, and the two groups have not reached a common level. The association is fragile. With starts as units, a post-hoc Welch interval for $D$ is $[-0.00001,+0.0022]$ ($p=0.053$ two-sided, $0.026$ one-sided); per seed, $D=0.0016\pm0.0006$ (seed 60) and $0.0005\pm0.0009$ (seed 61); leaving out one start at a time keeps $D$ positive in all twelve cases but $D-2\,\mathrm{se}>0$ in only four (\texttt{ms300\_sensitivity.py}). Two seeds shared across starts are not 24 independent replications. The walk-visited group's compatibility with the tilt does not show that the tilt is the stationary level, since chains of the same kernel from selected starts stand $0.0010$ lower. In the descriptive Haiku arm the walk-visited starts end at $0.9528\pm0.0020$ and the search-selected starts at $0.9535\pm0.0010$, both within two standard errors of the prediction ($0.9531$), with $D=-0.0003\pm0.0020$.

\subsection{Second domain: minibatch-gated prompt optimisation (test 5)}\label{sec:prompt}
Tests~2--4 use bin packing only. To check whether the inheritance relation of test~2 transfers, we registered a test in a second domain before its calibration began, with the decision rules, the evaluation script and the transferred line $\hat L(\rho)=1.18-0.52\,\rho$ (the six-setting bin-packing fit, residual s.d.\ $0.065$) fixed in advance.

\paragraph{Design.} The candidates are system prompts for a synthetic counting task: each instance asks how many times two given letters occur in total in a list of 8--12 words, and the answer is scored by exact match. A fixed set of $N=100$ instances defines the score vector, the 0/1 correctness of each instance, so true quality is the accuracy on all 100. Each prompt is evaluated once: the task model's answers are sampled once and the score vector is stored and reused, so true quality here is the \emph{frozen} empirical accuracy of that evaluation, not the expected accuracy of the prompt under re-sampling of the task model. The gates re-draw minibatches from the frozen vector (this is regime R, not C), but a search can select lucky evaluations as well as better prompts, and the design cannot separate the two. We did not re-evaluate prompts; the levels below are frozen-score levels, not reproducible model accuracies. The task model is Claude Haiku~5.5, answering 20 instances per call with the candidate as its system prompt. The proposer is also Haiku~5.5, starting from the prompt \emph{``Answer the questions.''} (frozen quality $0.50$ and $0.48$ in its two independent evaluations, one per setting). Two settings differ only in the instruction: \emph{p\_small}, the small-edit instruction of the bin-packing settings, and \emph{p\_bold}, \emph{``Write a substantially different prompt built on a new idea.''} The gate is regime R as before, with minibatches drawn without replacement from the 100 instances, and ties reject. Calibration used 6 no-selection walks of 60 steps per setting, with $\pi_0$ estimated from the pooled second halves (186 prompts). The prediction, $\tau$ and $\rho$ were computed by the bin-packing scripts and committed before search. Search used $m\in\{4,16,64\}$, 6 seeds and 100 steps; the observed level is the mean over steps 60--100. \emph{H7} (ordering, $m=64$ above $m=4$) is a sanity check and is not counted. \emph{H8} (direction): if $\rho$(p\_small) exceeds $\rho$(p\_bold) by at least $0.2$, then $L$(p\_bold) $>$ $L$(p\_small): pass if the difference minus two standard errors is positive, fail if the difference plus two standard errors is negative, inconclusive otherwise. \emph{H9} (transfer), per setting: pass if $|L-\hat L(\rho)|\le2\sqrt{0.065^2+\mathrm{se}^2}$. As in test~2, $L$ averages the cells with ESS at least $30$. An addendum, committed while the searches ran and before any search result existed, fixed four points: the standard error of $L$ is taken over seeds; only $m=4$ qualifies in either setting (ESS $129$, $29$, $3$ for p\_small and $81$, $18$, $3$ for p\_bold), so H8 and H9 rest on one cell per setting; a secondary prediction that rescales $\beta$ by the measured acceptance slope ($1.46$ and $1.45$ at $m=4$) is reported without a rule; and a task-model call that times out twice is scored as unparseable. The registered void conditions (spread of $\pi_0$ below $0.01$, or more than 10\% of evaluated prompts with an unparseable chunk) are not met: $\pi_0$ has s.d.\ $0.098$ and $0.134$, and $6.1\%$ and $2.8\%$ of prompts have an unparseable chunk.

\paragraph{Results.} Calibration gave $\rho=0.73$ (p\_small) and $0.29$ (p\_bold), a gap of $0.43$, so H8 is not void, and $\hat L=0.80$ and $1.02$. At $m=4$ the predictions are $0.474$ and $0.674$; search reached $0.756\pm0.075$ ($z=+3.5$) and $0.781\pm0.058$ ($z=+1.7$), so $L=5.98\pm1.32$ and $1.78\pm0.42$. These registered standard errors cover the search seeds only. A post-hoc bootstrap that also resamples the six calibration walks (through $\pi_0$, $\tau$, $\beta$ and $\rho$) gives 95\% intervals of $[3.0,20.6]$ and $[0.96,2.98]$ for $L$, and $[2.2,19.7]$ and $[-0.05,1.90]$ for $L-\hat L$ (\texttt{prompt\_L\_bootstrap.py}); it is conditional on the frozen scores and the fixed line, and six walks give no exact coverage. \emph{H8 fails}, in the opposite direction: $L$(p\_bold) $-$ $L$(p\_small) $=-4.20\pm1.38$ (bootstrap interval $[-18.6,-0.9]$). The setting with the higher parent--child correlation exceeds its prediction by more, whereas on bin packing the higher-correlation settings fell short. \emph{H9 fails for p\_small} ($|L-\hat L|=5.18$, tolerance $2.64$) \emph{and passes for p\_bold} ($0.75$, tolerance $0.85$); with the calibration uncertainty above, the pass means that the transferred line is not rejected there, not that it predicts accurately. H7 holds in both settings ($0.963$ against $0.756$; $0.903$ against $0.781$), although p\_bold is not monotone in $m$ ($0.916$ at $m=16$). The secondary, slope-corrected prediction is still exceeded ($L=3.85$ and $1.19$). In five of the six cells, including the four with ESS below $30$, search lies above the prediction. In this domain, therefore, the tilt computed from the short calibration walks was a \emph{lower} estimate of what search reached in 100 steps on the frozen scores, and the finite-step calibration and prediction procedure that worked directionally on bin packing did not transfer; we do not read this as isolating an inheritance mechanism that reverses across domains.

\paragraph{A post-hoc observation.} The p\_small calibration walks are not stationary. Their mean quality falls from $0.61$ over the first ten steps to $0.41$ by step 40 and is $0.45$ over the last ten, so unselected small edits degrade the prompt, and the second-half sample, which estimates $\pi_0$, has maximum $0.73$. A reweighting of that sample cannot exceed its maximum, and the search level at $m=4$ ($0.756$) is above it: the prediction is capped by the region the walks occupied when the sample was taken, not by the gate. Earlier steps of the same walks reached $0.96$. On bin packing the walks were longer (100 steps) and the search started near the walk mean. That short calibration walks cause the underestimate here is a reading after the result. We did not test it, and it is consistent with the general caution of Section~\ref{sec:llm} that a calibration walk may not sample the kernel's equilibrium near the top.

\subsection{Exploratory run: cached versus re-evaluated search}\label{sec:llm-explore}
Before the pre-registered tests we ran a single exploratory setting (BP, Haiku, neutral instruction) with both regimes. Its prediction is not pre-registered: $\tau$ was estimated mostly from search pairs, without the finite-population factor, and the walk sample was small (6 walks, 306 programs). We report it for the comparison of regimes, which the pre-registered runs do not cover.

\paragraph{Design.} Regime R as above. Regime C compares the child's minibatch score with the score cached when the incumbent was accepted. Both regimes run $m\in\{1,4,16,64\}$ with 6 seeds, for 100 steps ($m\le4$) or 200 steps ($m\ge16$); within a seed, the $k$-th proposal from a given parent is shared across regimes and minibatch sizes (common random numbers). In total 8{,}927 programs were generated; 7 failed to run.

\paragraph{Inputs and predictors.} The mutation equilibrium $\pi_0$ comes from 6 no-selection walks of 100 steps, second halves pooled, 306 points with mean $0.9423$, s.d.\ $0.0077$ and maximum $0.9550$. The per-instance difference scale $\tau=0.0061$ is the median, over 8{,}607 parent--child pairs, of the standard deviation of per-instance score differences; most of those pairs come from the search runs, and using only the 547 walk pairs gives $\tau=0.0054$ and moves every tilt prediction by at most $0.0004$. The \emph{tilt prediction} plugs $\beta(m)=1.596\sqrt m/\tau$ into Corollary~\ref{cor:plateau}. The \emph{projected-kernel predictor} drops reversibility: it estimates the conditional law of child quality given parent quality from all generated pairs, including the search's own, and evaluates the one-dimensional chain with acceptance $\Phi(\Delta\sqrt m/\tau)$; it is a consistency check rather than an independent prediction.

\begin{table}[h]\centering
\caption{Exploratory run (not pre-registered): predicted vs.\ observed level, regimes R and C. Values are true quality of the incumbent, averaged over the last third of the run (100 steps for $m\le4$, 200 for $m\ge16$). $\pm$: standard error over 6 seeds, or bootstrap error over walks for the tilt prediction. ESS: effective sample size of the reweighting.}\label{tab:llm}
\begin{tabular}{rccrccc}\toprule
$m$ & $\beta(m)$ & tilt pred. & ESS & kernel & R observed & C observed\\\midrule
1 & 260 & $0.9502\pm.0004$ & 129 & 0.9497 & $0.9505\pm.0011$ & $0.9503\pm.0008$\\
4 & 519 & $0.9515\pm.0004$ & 58 & 0.9514 & $0.9509\pm.0009$ & $0.9508\pm.0008$\\
16 & 1038 & $0.9533\pm.0006$ & 10 & 0.9540 & $0.9535\pm.0008$ & $0.9521\pm.0008$\\
64 & 2076 & $0.9545\pm.0006$ & 2.6 & 0.9563 & $0.9551\pm.0002$ & $0.9544\pm.0002$\\\bottomrule
\end{tabular}\end{table}

\begin{figure}[!htbp]\centering\includegraphics[width=\linewidth]{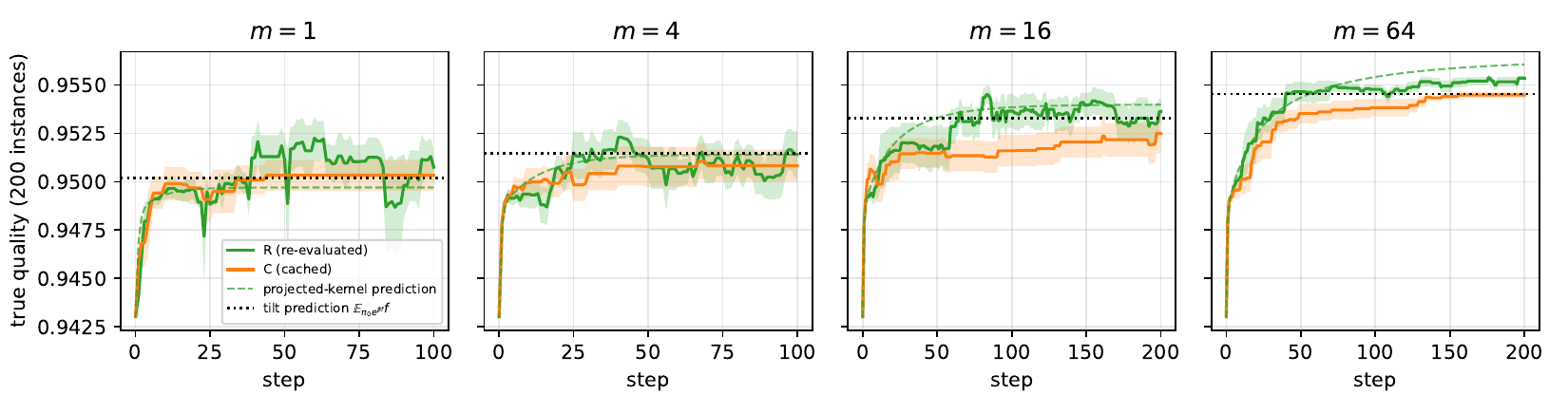}
\caption{Exploratory run, mean $\pm$ s.e.\ over 6 seeds. Solid lines and bands: observed true quality of the incumbent. Dotted: tilt prediction from $\pi_0$ and $\tau$ alone. Dashed: projected-kernel transient.}\label{fig:llm}\end{figure}

\paragraph{Results.} Table~\ref{tab:llm} and Figure~\ref{fig:llm}. Regime R levels off, and the level rises with $m$ from $m=4$ on. The tilt prediction differs from the observed level by $+0.0003$, $-0.0006$, $+0.0002$ and $+0.0006$ for $m=1,4,16,64$, each within one combined standard error; at $m\ge16$ the ESS is $10$ and $3$, so this agreement is weak evidence, and with the larger walk samples and walk-only $\tau$ of Section~\ref{sec:llm} the same setting is predicted $0.0006$--$0.0010$ higher, above the fresh runs of test~1. At matched steps, R is above C at $m=64$ ($+0.0007\pm0.0002$); at $m=16$ the difference ($+0.0014\pm0.0008$) is $1.75$ standard errors, and at $m\le4$ the two are level.

\paragraph{Cached search inflates its own score.} In regime C we replay each chain and compare the recorded score of the final incumbent with its true quality. The recorded score is too high by $0.0151\pm0.0015$ ($m=1$), $0.0099\pm0.0012$ ($m=4$), $0.0033\pm0.0007$ ($m=16$) and $0.0015\pm0.0001$ ($m=64$); at $m=1$ the overstatement is three times the entire difference in true quality between $m=1$ and $m=64$. At matched \emph{evaluations} (C after $2t$ steps against R after $t$ steps), C is not ahead within our budget: $C-R=-0.0008\pm0.0017$, $-0.0003\pm0.0014$, $-0.0012\pm0.0007$, $-0.0003\pm0.0004$ for $m=1,4,16,64$. In the simulations of Section~\ref{sec:sim} C is ahead of R at every budget, and neither bounded quality nor larger steps changes that (\texttt{sim\_cache\_bounded.py}). Pairing is a plausible reason for the difference. R scores parent and child on the same minibatch, so the difficulty of the sampled instances cancels; C compares against a score from a different minibatch. In our data the unpaired comparison has s.d.\ $\sqrt2\times0.0082=0.0116$ against $\tau=0.0062$ for the paired one. Giving C $1.87$ times R's per-evaluation noise in the simulation makes C level with or behind R up to about 500 evaluations in most settings, and ahead only at larger budgets (\texttt{sim\_cache\_paired.py}); our LLM runs use at most 400. We have not run the direct contrast (R with unpaired minibatches). Once the recorded score is inflated, almost no proposal clears it: in the second half of the run, C chains accept $0.0$, $0.3$, $1.5$ and $1.3$ proposals per seed for $m=1,4,16,64$, against $7.8$, $10.3$, $13.0$ and $5.5$ for R.

\paragraph{Caveats.} The LLM mutation is not known to be reversible and its steps are large, so Theorem~\ref{thm:smallstep} does not formally apply; the experiments measure how well the small-step temperature transfers rather than assume it. A pilot reused a fixed pool of 8 children per parent. It stalled at parents with no better child and gave predictions $0.002$--$0.004$ too high. Fresh proposals removed most of that gap, which suggests the prediction applies when the mutation kernel is sampled, not when it is enumerated from a small cache. All programs, scores and scripts are archived with the paper (Appendix~\ref{app:llm}).

\section{Related work}\label{sec:related}
\paragraph{Noisy acceptance and tilted stationary laws.} Logistic choice from Gumbel utility shocks is due to \citet{mcfadden1974} and \citet{yellott1977}; Barker's rule to \citet{barker1965}; its equivalence to ``accept iff $\Delta$ plus logistic noise is positive'' is used by \citet{seita2017}. Logit dynamics yield Gibbs stationary laws in potential games \citep{blume1993,alosferrer2010}. In population genetics, the stationary law under weak mutation is the mutation equilibrium tilted by fitness with population size as inverse temperature \citep{iwasa1988,sella2005}; \citet{wu2010} show that, under weak selection, fixation probabilities and fixation times of pairwise-comparison processes depend on the comparison function only through its slope at zero. Theorem~\ref{thm:smallstep} is a stationary-law analogue in a diffusion limit; we are not aware of a prior statement of $\beta=4F'(0)$ for the stationary law of keep-if-better search, but the computation is elementary. Closest in spirit, \citet{traulsen2007stochastic} show for evolutionary games that stochastic payoff evaluation lowers the intensity of selection, i.e.\ raises the temperature of a Fermi pairwise-comparison process \citep{traulsen2007pairwise}, via a mean-field effective temperature, and study fixation. Our title restates their observation for LLM-guided search; what we add is the stationary law of keep-if-better search with a mutation operator, the constant $4F'(0)$ for general symmetric noise, the paired-minibatch form, and a stationary level that, under the stated conditions, can be computed from unselected mutation before search (whether finite runs reach it is tested in Section~\ref{sec:llm}). In evolutionary computation, Boltzmann tournament selection builds a Boltzmann distribution from pairwise competitions \citep{goldberg1990boltzmann}, and \citet{rattray1998noisy} analyse Boltzmann selection under additive Gaussian fitness noise, where in the weak-selection limit noise can be offset by a larger population. Noisy MCMC treats evaluation noise as bias to be corrected \citep{ceperley1999,alquier2016}; we treat it as the temperature the search actually runs at.
\paragraph{Noisy evolutionary optimisation.} The role of re-evaluating the parent is analysed for evolution strategies on the noisy sphere by \citet{arnold2002}, with closed-form residual distances in the steady state \citep{beyer2001}; runtime theory for discrete noisy optimisation is surveyed by \citet{jin2005}. These results concern specific landscapes and step-size adaptation rather than the stationary law of a generic mutation operator.
\paragraph{LLM-guided search and verifiers.} SMCEvolve \citep{jiang2026smcevolve} targets the same tilted law $p_0e^{\beta R}$, with the terminal $\beta$ set as a parameter, an adaptive (ESS-based) schedule of intermediate temperatures, and an exact evaluator. Here the temperature is not chosen but induced by the verifier noise. \citet{hu2026winner} study the winner's curse of keep-if-better prompt optimisation empirically. In one exploratory setting with three seeds, re-drawing the selection set lowered the final held-out gain; they note that this comparison is confounded with feedback and with the effective number of candidates, so we cite it as related evidence rather than as a test of our results. For independent sampling with imperfect verifiers, \citet{dorner2025roc} and \citet{stroebl2024} derive ceilings from false-positive rates and ROC slopes, and \citet{kwa2024} characterise heavy-tailed reward error. Empirical comparisons of evolutionary and independent search at matched budgets \citep{oved2026illusion,oved2026tokens} motivate the question but do not model verifier noise; most of their evaluators are deterministic.

\section{Discussion and limitations}
\paragraph{What to do with a noisy verifier.} The results suggest three measurements for a re-evaluated loop, all of which need no search: estimate $\beta$ from the per-task score differences of a sample of parent--child pairs, checking the acceptance rule directly when $m$ is small or scores are discrete; estimate the quality distribution of unselected mutation; and compute the tilt $\kappa'(\beta)$ (strictly, the stationary mean of the tilted law, which Theorem~\ref{thm:smallstep} establishes only for the small-step limit). A larger minibatch raises $\beta\propto\sqrt m$ and, in six of our seven settings, raised the level the search reached. Whether more steps will help is a different question, which the tilt does not answer for LLM editing: in our runs, chains from different starts had not met after 150--300 steps, so a single run that has flattened is not evidence that the loop is at its stationary level. In our bin-packing experiments search attained $62$--$88\%$ of the predicted gain over unselected mutation after 150 steps with proposers that edit the parent ($67$--$93\%$ after 300 steps in the four settings continued) and $109$--$119\%$ with proposers that write new heuristics. On bin packing the parent--child quality correlation of the calibration walk, which costs nothing extra, indicated which case applied: for an editing proposer the prediction was an upper estimate of what search from the root reached within 300 steps, and the fitted line of Section~\ref{sec:llm} gives a rough correction, whose quantitative accuracy we could not establish. Neither held in the prompt domain (Section~\ref{sec:prompt}): search exceeded the tilt in both settings, more so with the small-edit instruction, and the calibration walks there were still drifting. The correction should not be used outside the task it was fitted on, and a calibration walk should be checked for drift before its tilt is trusted as an estimate in either direction.

\paragraph{Limitations.} The theory assumes a mutation kernel reversible with respect to its equilibrium; whether LLM mutation is reversible is unknown, and our experiments measure rather than assume how well the prediction transfers. For non-logistic noise the result is a small-step diffusion approximation, and we do not prove convergence of the stationary laws; we document a large-step regime where mixing becomes very slow, and leave existence in general open. We give no growth law for cached search and no mixing-time bound, so the \emph{early} crossover between independent and re-evaluated search is measured, not predicted.
The program-search study covers two tasks, two proposer models of one family and three mutation instructions, each with six seeds and 150 steps (300 for four settings), and the prompt test two settings with one model, 100 steps and calibration walks of only 60 steps; on the travelling-salesman task the seed-to-seed spread is too large for a sharp test, and at the largest minibatch sizes the walk samples are too small to estimate the tilt (effective sample sizes in single digits). All reported levels are levels of finite runs; none of our tests establishes a stationary level. The calibration walks themselves may not be stationary: in one setting their mean is still drifting over the second half, by an amount that moves the prediction as much as the shortfall (Appendix~\ref{app:diag}).
The inheritance account is an empirical regularity on one task with one model family: the linear relation was found on four settings, its direction confirmed on two, its quantitative form is not robust to how the uncertainty of the registered test is computed, and in a second domain its registered direction failed and its quantitative form was not rejected in one of two settings, with wide calibration uncertainty. Our post-hoc reading of that failure (short calibration walks that cap the prediction) was not tested. Two registered analyses had flaws that we found after the results: the standard error of the attained fraction ignored proposals shared across minibatch sizes, and the selection-effect statistic pooled proposals in a way that over-weights hard-to-improve search parents. We report the registered verdicts and the corrected ones side by side. The explanations of the residual errors are post-hoc. For proposers whose children are nearly independent of the parent's quality, the tilt is too low because quality depends on lineage depth and the calibration walks are deeper than the search; a calibration-only correction removes the error but was found on the same data, and the overshoot should shrink with run length, which we did not test. For editing proposers, a reversible null model with the walks' correlation structure does not reproduce the runs from above, and a registered 300-step test confirms that these runs do not recover. A fourth registered test associates this with the start's origin, in one setting at $m=4$ and among twelve starts whose origin was not randomised: from programs of the same quality that unselected walks visited, the runs kept their level for 150 steps (secondary window; the primary window was inconclusive) and were statistically compatible with the prediction after 300 steps, while runs from programs that searches at $m=16$ and $64$ had selected lost it and stayed below the prediction, apart from the first group by a margin that passed the registered rule narrowly and is sensitive to leaving out single starts. In the descriptive second setting both groups lost quality. Our data therefore do not determine whether the stationary level of the small-edit kernel lies below the tilt, at it, or whether a common level exists within practical run lengths; a kernel that is reversible with respect to $\pi_0$ but metastable is not excluded. In one cell (Sonnet, rewrite, $m=4$) the incumbent is still falling at 150 steps. The Gaussian approximation to the acceptance rule is too flat at the smallest minibatch, where scores are discrete and ties are frequent, and the per-pair noise scale is heterogeneous, so $\beta$ is a summary rather than a constant.

\bibliographystyle{plainnat}
\bibliography{refs}

\begin{thebibliography}{32}
\providecommand{\natexlab}[1]{#1}
\providecommand{\url}[1]{\texttt{#1}}
\expandafter\ifx\csname urlstyle\endcsname\relax
  \providecommand{\doi}[1]{doi: #1}\else
  \providecommand{\doi}{doi: \begingroup \urlstyle{rm}\Url}\fi

\bibitem[Agrawal et~al.(2025)]{agrawal2025gepa}
Lakshya~A. Agrawal et~al.
\newblock {GEPA}: Reflective prompt evolution can outperform reinforcement
  learning.
\newblock \emph{arXiv:2507.19457}, 2025.

\bibitem[Al{\'o}s-Ferrer and Netzer(2010)]{alosferrer2010}
Carlos Al{\'o}s-Ferrer and Nick Netzer.
\newblock The logit-response dynamics.
\newblock \emph{Games and Economic Behavior}, 68:\penalty0 413--427, 2010.

\bibitem[Alquier et~al.(2016)Alquier, Friel, Everitt, and Boland]{alquier2016}
Pierre Alquier, Nial Friel, Richard Everitt, and Aidan Boland.
\newblock Noisy {Monte Carlo}: convergence of {Markov} chains with approximate
  transition kernels.
\newblock \emph{Statistics and Computing}, 26:\penalty0 29--47, 2016.

\bibitem[Arnold and Beyer(2002)]{arnold2002}
Dirk~V. Arnold and Hans-Georg Beyer.
\newblock Local performance of the (1+1)-{ES} in a noisy environment.
\newblock \emph{IEEE Transactions on Evolutionary Computation}, 6\penalty0
  (1):\penalty0 30--41, 2002.

\bibitem[Barker(1965)]{barker1965}
A.~A. Barker.
\newblock Monte {Carlo} calculations of the radial distribution functions for a
  proton-electron plasma.
\newblock \emph{Australian Journal of Physics}, 18:\penalty0 119--133, 1965.

\bibitem[Beyer(2001)]{beyer2001}
Hans-Georg Beyer.
\newblock \emph{The Theory of Evolution Strategies}.
\newblock Springer, 2001.

\bibitem[Blume(1993)]{blume1993}
Lawrence~E. Blume.
\newblock The statistical mechanics of strategic interaction.
\newblock \emph{Games and Economic Behavior}, 5:\penalty0 387--424, 1993.

\bibitem[Ceperley and Dewing(1999)]{ceperley1999}
D.~M. Ceperley and M.~Dewing.
\newblock The penalty method for random walks with uncertain energies.
\newblock \emph{Journal of Chemical Physics}, 110:\penalty0 9812, 1999.

\bibitem[Dorner et~al.(2026)Dorner, Chen, Cruz, and Yang]{dorner2025roc}
Florian~E. Dorner, Yatong Chen, Andr{\'e}~F. Cruz, and Fanny Yang.
\newblock {ROC-n-reroll}: How verifier imperfection affects test-time scaling.
\newblock In \emph{International Conference on Learning Representations
  (ICLR)}, 2026.
\newblock arXiv:2507.12399.

\bibitem[Ethier and Kurtz(1986)]{ethier1986}
Stewart~N. Ethier and Thomas~G. Kurtz.
\newblock \emph{Markov Processes: Characterization and Convergence}.
\newblock Wiley, 1986.

\bibitem[Goldberg(1990)]{goldberg1990boltzmann}
David~E. Goldberg.
\newblock A note on {B}oltzmann tournament selection for genetic algorithms and
  population-oriented simulated annealing.
\newblock \emph{Complex Systems}, 4\penalty0 (4):\penalty0 445--460, 1990.

\bibitem[Hu and Tang(2026)]{hu2026winner}
Litao Hu and Yutong Tang.
\newblock The winner's curse in {LLM} self-improvement loops: selection noise,
  lock-in, and acceptance rules.
\newblock \emph{arXiv:2610.09239}, 2026.

\bibitem[Iwasa(1988)]{iwasa1988}
Yoh Iwasa.
\newblock Free fitness that always increases in evolution.
\newblock \emph{Journal of Theoretical Biology}, 135:\penalty0 265--281, 1988.

\bibitem[Jiang et~al.(2026)Jiang, Zhu, and Zhu]{jiang2026smcevolve}
Jiachen Jiang, Huminhao Zhu, and Zhihui Zhu.
\newblock {SMCEvolve}: Principled scientific discovery via sequential {Monte
  Carlo} evolution.
\newblock \emph{arXiv:2605.15308}, 2026.

\bibitem[Jin and Branke(2005)]{jin2005}
Yaochu Jin and J{\"u}rgen Branke.
\newblock Evolutionary optimization in uncertain environments---a survey.
\newblock \emph{IEEE Transactions on Evolutionary Computation}, 9\penalty0
  (3):\penalty0 303--317, 2005.

\bibitem[Kwa et~al.(2024)Kwa, Thomas, and Garriga-Alonso]{kwa2024}
Thomas Kwa, Drake Thomas, and Adri{\`a} Garriga-Alonso.
\newblock Catastrophic {Goodhart}: regularizing {RLHF} with {KL} divergence
  does not mitigate heavy-tailed reward misspecification.
\newblock \emph{arXiv:2407.14503}, 2024.

\bibitem[McFadden(1974)]{mcfadden1974}
Daniel McFadden.
\newblock Conditional logit analysis of qualitative choice behavior.
\newblock In P.~Zarembka, editor, \emph{Frontiers in Econometrics}, pages
  105--142. Academic Press, 1974.

\bibitem[Novikov et~al.(2025)]{novikov2025}
Alexander Novikov et~al.
\newblock {AlphaEvolve}: A coding agent for scientific and algorithmic
  discovery.
\newblock \emph{arXiv:2506.13131}, 2025.

\bibitem[Opsahl-Ong et~al.(2024)]{opsahl2024}
Krista Opsahl-Ong et~al.
\newblock Optimizing instructions and demonstrations for multi-stage language
  model programs.
\newblock In \emph{Proceedings of EMNLP}, 2024.

\bibitem[Oved et~al.(2026{\natexlab{a}})Oved, Pony, Naparstek, and
  Barzelay]{oved2026illusion}
Tal Oved, Roi Pony, Oshri Naparstek, and Udi Barzelay.
\newblock Evolution or illusion? {Rethinking} evaluation in {LLM} evolutionary
  search.
\newblock \emph{arXiv:2609.19799}, 2026{\natexlab{a}}.

\bibitem[Oved et~al.(2026{\natexlab{b}})Oved, Pony, Naparstek, and
  Barzelay]{oved2026tokens}
Tal Oved, Roi Pony, Oshri Naparstek, and Udi Barzelay.
\newblock {LLM} evolutionary search strategies should be compared by tokens.
\newblock \emph{arXiv:2610.23977}, 2026{\natexlab{b}}.

\bibitem[Rattray and Shapiro(1997)]{rattray1998noisy}
Magnus Rattray and Jon Shapiro.
\newblock Noisy fitness evaluation in genetic algorithms and the dynamics of
  learning.
\newblock In Richard~K. Belew and Michael~D. Vose, editors, \emph{Foundations
  of Genetic Algorithms 4}, pages 117--139. Morgan Kaufmann, 1997.

\bibitem[Romera-Paredes et~al.(2024)]{romera2024}
Bernardino Romera-Paredes et~al.
\newblock Mathematical discoveries from program search with large language
  models.
\newblock \emph{Nature}, 625:\penalty0 468--475, 2024.

\bibitem[Seita et~al.(2017)Seita, Pan, Chen, and Canny]{seita2017}
Daniel Seita, Xinlei Pan, Haoyu Chen, and John Canny.
\newblock An efficient minibatch acceptance test for {Metropolis-Hastings}.
\newblock In \emph{Proceedings of UAI}, 2017.

\bibitem[Sella and Hirsh(2005)]{sella2005}
Guy Sella and Aaron~E. Hirsh.
\newblock The application of statistical physics to evolutionary biology.
\newblock \emph{Proceedings of the National Academy of Sciences}, 102\penalty0
  (27):\penalty0 9541--9546, 2005.

\bibitem[Smith and Winkler(2006)]{smith2006}
James~E. Smith and Robert~L. Winkler.
\newblock The optimizer's curse: skepticism and postdecision surprise in
  decision analysis.
\newblock \emph{Management Science}, 52\penalty0 (3):\penalty0 311--322, 2006.

\bibitem[Stroebl et~al.(2024)Stroebl, Kapoor, and Narayanan]{stroebl2024}
Benedikt Stroebl, Sayash Kapoor, and Arvind Narayanan.
\newblock Inference scaling {fLaws}: the limits of {LLM} resampling with
  imperfect verifiers.
\newblock \emph{arXiv:2411.17501}, 2024.

\bibitem[Traulsen et~al.(2007{\natexlab{a}})Traulsen, Nowak, and
  Pacheco]{traulsen2007stochastic}
Arne Traulsen, Martin~A. Nowak, and Jorge~M. Pacheco.
\newblock Stochastic payoff evaluation increases the temperature of selection.
\newblock \emph{Journal of Theoretical Biology}, 244\penalty0 (2):\penalty0
  349--356, 2007{\natexlab{a}}.

\bibitem[Traulsen et~al.(2007{\natexlab{b}})Traulsen, Pacheco, and
  Nowak]{traulsen2007pairwise}
Arne Traulsen, Jorge~M. Pacheco, and Martin~A. Nowak.
\newblock Pairwise comparison and selection temperature in evolutionary game
  dynamics.
\newblock \emph{Journal of Theoretical Biology}, 246\penalty0 (3):\penalty0
  522--529, 2007{\natexlab{b}}.

\bibitem[Verdun et~al.(2025)Verdun, Oesterling, Lakkaraju, and
  Calmon]{verdun2025soft}
Claudio~Mayrink Verdun, Alex Oesterling, Himabindu Lakkaraju, and Flavio~P.
  Calmon.
\newblock Soft best-of-n sampling for model alignment.
\newblock In \emph{IEEE International Symposium on Information Theory (ISIT)},
  2025.
\newblock arXiv:2505.03156.

\bibitem[Wu et~al.(2010)Wu, Altrock, Wang, and Traulsen]{wu2010}
Bin Wu, Philipp~M. Altrock, Long Wang, and Arne Traulsen.
\newblock Universality of weak selection.
\newblock \emph{Physical Review E}, 82:\penalty0 046106, 2010.

\bibitem[Yellott(1977)]{yellott1977}
John~I. Yellott.
\newblock The relationship between {Luce's} choice axiom, {Thurstone's} theory
  of comparative judgment, and the double exponential distribution.
\newblock \emph{Journal of Mathematical Psychology}, 15:\penalty0 109--144,
  1977.

\end{thebibliography}
\appendix
\section{Proofs}\label{app:proofs}
\paragraph{Theorem~\ref{thm:smallstep}.} Let $g$ be smooth with compact support and write $\Delta_h=f(y)-f(x)$, $y\sim K_h(x,\cdot)$. The generator of regime R is $L_hg(x)=\E\big[(g(y)-g(x))F(\Delta_h)\big]$. Since $F(\Delta)+F(-\Delta)=1$ and $F$ is differentiable at $0$, $F(\Delta)=\tfrac12+F'(0)\Delta+r(\Delta)$ with $r(\Delta)=o(\Delta)$. Then
\[L_hg=\tfrac12(K_h-I)g+F'(0)\,\E[(g(y)-g(x))(f(y)-f(x))]+\E[(g(y)-g(x))\,r(\Delta_h)].\]
By assumption $h^{-2}\tfrac12(K_h-I)g\to\tfrac14(\Delta g+\nabla\log\pi_0\cdot\nabla g)$. For $x$ in a compact set, where the moment bounds hold uniformly, Taylor expansion with the assumed step covariance $h^2I+o(h^2)$ and third moment $O(h^3)$ gives $\E[(g(y)-g(x))(f(y)-f(x))]=h^2\nabla g\cdot\nabla f+o(h^2)$. For the remainder, write $\xi=y-x$ and use $|\Delta_h|\le L|\xi|$ and $|g(y)-g(x)|\le L|\xi|$. Given $\eta>0$ choose $\epsilon$ with $|r(\Delta)|\le\eta|\Delta|$ for $|\Delta|<\epsilon$. On that event the remainder is at most $\eta L^2\E|\xi|^2=O(\eta h^2)$. On the complement, $|r(\Delta)|\le 1+F'(0)|\Delta|$ and Markov's inequality give $O(\E|\xi|^3/\epsilon^2)=O(h^3)$. Hence the remainder is $o(h^2)$. Hence $h^{-2}L_hg\to\mathcal Lg$ uniformly on compact sets. The operator $\mathcal L=\tfrac14(\Delta+\nabla U\cdot\nabla)$ with $U=\log\pi_0+\beta f$ is the generator of the Langevin diffusion $\mathrm dX=\tfrac14\nabla U\,\mathrm dt+\tfrac1{\sqrt2}\mathrm dW$, which is reversible with respect to $e^{U}$ whenever $e^U$ is integrable. Convergence of the rescaled chains to this diffusion on compact time intervals follows from standard generator-convergence results for Markov chains \citep[Ch.~4, Sec.~8]{ethier1986} under the usual tightness conditions; we do not claim convergence of the stationary laws themselves. That would need, for example, tightness of the chains' invariant laws, uniqueness of the invariant law of the limit, and uniform integrability of $f$; uniform-in-$h$ ergodicity is one sufficient route. Section~3.4 shows that mixing can become very slow for large steps, so such conditions are not automatic. Accordingly, we use Theorem~\ref{thm:smallstep} as a diffusion approximation, and its use for LLM search (Section~\ref{sec:llm}) is an empirical extrapolation.

\paragraph{Proposition~\ref{prop:bon}(a), Gaussian quality.} Let $f_1,\dots,f_N$ be i.i.d.\ $N(\mu,s^2)$, independent of i.i.d.\ Gumbel$(0,b)$ noise, $\beta=1/b$, $I_N=\arg\max_i(f_i+\varepsilon_i)$ and $p_i=e^{\beta f_i}/\sum_je^{\beta f_j}$. By the Gumbel-max property, $\Pr(I_N=i\mid f_1,\dots,f_N)=p_i$, so $\E f_{I_N}=\sum_i\E[f_ip_i]$. Since $0\le p_i\le1$ and $\partial p_i/\partial f_i=\beta p_i(1-p_i)$ is bounded, Gaussian integration by parts (Stein's identity) gives $\E[(f_i-\mu)p_i]=s^2\E[\partial p_i/\partial f_i]=\beta s^2\E[p_i(1-p_i)]$, and summing over $i$, $\E f_{I_N}=\mu+\beta s^2\big(1-\E\sum_ip_i^2\big)$. With $W_i=e^{\beta f_i}$, which has finite first and second moments, $\sum_ip_i^2=N^{-1}\big(N^{-1}\sum_iW_i^2\big)/\big(N^{-1}\sum_iW_i\big)^2\to0$ almost surely by the strong law, and since $0\le\sum_ip_i^2\le1$, also in mean. Hence $\E f_{I_N}\to\mu+\beta s^2=\kappa'_{\mathrm{root}}(\beta)$. This covers Gaussian quality with Gumbel noise only; it is distinct from part (b).

\paragraph{Gaussian constants.} With i.i.d.\ $N(0,\sigma^2)$ noise on each evaluation, the difference has s.d.\ $\sqrt2\sigma$, $F'(0)=1/(\sqrt{2\pi}\sqrt2\sigma)=1/(2\sqrt\pi\sigma)$, and $\beta=2/(\sqrt\pi\sigma)$. With a paired minibatch, the difference has s.d.\ $\tau/\sqrt m$ directly, and $\beta=4\sqrt m/(\sqrt{2\pi}\tau)$.

\section{Simulation details}\label{app:sims}
All simulations use the autoregressive mutation of Example~\ref{ex:ar} with quality as the state, $\pi_{\mathrm{root}}=N(0,1)$, and score noise of s.d.\ $1/d$ on each evaluation, so $d$ is the signal-to-noise ratio of a single score relative to the spread of fresh samples. Code is in the supplementary repository (\texttt{sims/}).
\begin{itemize}
\item \texttt{sim\_tilt.py}: Gumbel and Gaussian noise, 20{,}000 parallel chains, 3{,}000 steps.
\item \texttt{sim\_beta\_limit.py}: mean divided by $d$ for $\rho\in\{0.5,0.9,0.99,0.998\}$; runs of $60/(1-\rho)+500$ steps, averaged over the second half.
\item \texttt{predict\_plateau.py}: the stationary law of regime R computed on a 1{,}500-point grid by power iteration on the discretised transition matrix, started from the mutation equilibrium; no search is simulated.
\item \texttt{sim\_matched\_evals.py}: IID, C and R at a matched number $E$ of evaluations.
\item \texttt{sim\_inflation.py}: recorded minus true quality in regime C.
\item \texttt{check\_nostat.py}: long-run averages for the slow-mixing example of Section~3.4.
\item \texttt{sim\_cache\_bounded.py}, \texttt{sim\_cache\_paired.py}: regime C against R with bounded quality, small $\rho$, and unpaired noise for C.
\end{itemize}

\section{LLM experiment details}\label{app:llm}
\paragraph{Proposer calls.} Each mutation is a single-turn call with no tools, default sampling parameters, a system prompt describing the task, and a user prompt consisting of the current program followed by an instruction and an interface statement. A reply's last fenced code block is taken as the child; if it does not define \texttt{priority}, the call is repeated up to three times, after which a program that raises is recorded (it is never accepted). Transport errors are retried with back-off and never recorded as children. Calls go through a command-line client with the default system prompt replaced, tools disabled and one turn; from 2026-11-03 20:50 UTC the runs still in progress (test~4's continuation and the second-domain searches) were moved to another machine and billed to an API key instead of a subscription, with the same client, models and flags. Stored children replay identically on either machine. \emph{Restarts.} Every child is written only after it is generated and evaluated, and every restart used the same command, so it replayed the stored steps and generated no child twice. Besides the Sonnet multi-start restart described below, the runs were restarted when they moved machine (20:50 UTC) and 19 minutes later, when the spend ledger was corrected to price calls from per-model usage. The two prompt-domain searches were also restarted after one crash each, caused by an uncaught task-model timeout (the timeout rule of the addendum was added then, before any search result existed), when the spend cap was raised (2026-11-04 04:04 UTC), and after a pause for a spend check (resumed 16:37 UTC). None of these restarts changed a model, prompt, rule or seed. Logs are appended with a marker line at each restart. Children are cached by parent and proposal index, so a proposal shared across regimes or minibatch sizes is the same program.
\paragraph{Prompts.} System prompt (BP): \emph{``You write Python heuristics for online bin packing. Bins have capacity 100; items arrive one at a time and each goes to the feasible bin with the highest priority (a new bin is opened only if none fits). Objective: use as few bins as possible. Item sizes follow a Weibull-like distribution, typically 20-70.''} System prompt (TSP): \emph{``You write Python heuristics for constructing short travelling-salesman tours. A tour starts at city 0; at each step the next city is the unvisited city with the highest priority. Cities are 60 random points in the unit square. Objective: the shortest closed tour.''} User prompt: \emph{``Here is the current priority function:''}, the program, then one of the instructions
neutral: \emph{``Write an improved or alternative version.''};
small edit: \emph{``Make one small, targeted change that you expect to improve it; keep everything else identical.''};
rewrite: \emph{``Write a substantially different heuristic built on a new idea, not a tweak of this one.''};
followed by the interface statement. System prompt (prompt domain): \emph{``You write instructions (a system prompt) for a language model that answers batches of letter-counting questions: how many times a given letter appears in total in a short list of words. Each final answer is scored by exact match. Objective: the highest fraction of correct answers.''} (the instances ask about two letters combined); user prompt \emph{``Here is the current prompt:''}, the prompt, the p\_small or p\_bold instruction, and \emph{``The questions and the required answer format are appended automatically, so the prompt only needs to guide how the model works. Keep it under 120 words. Reply with only the new prompt in one code block.''} Interface statements (BP: \emph{``Signature must stay \texttt{def priority(item, bins)}: \texttt{item} is a float, \texttt{bins} is a numpy array of remaining capacities of bins that can fit the item; return a numpy array of the same length (higher = preferred). \texttt{np} and \texttt{math} are available; no imports, no file or network access. Keep it under 40 lines. Reply with only one python code block.''}; TSP analogous, for \texttt{priority(cur, cand, dist, visited\_frac)}).
\paragraph{Evaluation.} Each program runs once, in an isolated interpreter with a 60\,s limit, on all 200 instances of its task; the per-instance scores are stored, and every minibatch score is computed from them. A program that raises or exceeds the limit has no score.
\paragraph{Programs generated.} BP/Haiku/neutral: 19{,}653 programs (both runs; 0.1\% failed). BP/Haiku/small edit: 10{,}480 (0.3\%). BP/Haiku/rewrite: 6{,}322 (7.3\%). BP/Sonnet/neutral: 11{,}792 (0.04\%). BP/Sonnet/small edit: 17{,}113 (0.01\%). BP/Sonnet/rewrite: 6{,}347 (0.9\%). TSP/Haiku/neutral: 6{,}475 (1.9\%). Prompt domain: 2{,}097 (p\_small) and 2{,}126 (p\_bold) prompts, of which $6.1\%$ and $2.8\%$ have at least one unparseable chunk. Identical programs are stored once. On calibration pairs, the child's whole score vector equals the parent's in $10.0\%$ (Haiku neutral), $45.3\%$ (Haiku small edit), $4.9\%$ (Sonnet neutral), $35.9\%$ (Sonnet small edit), $1.4\%$ and $1.0\%$ (rewrite) and $0.3\%$ (TSP) of pairs. Such a proposal can never pass a strict gate, so 150 steps offer the small-edit searches fewer score-changing proposals than the others. Without these pairs $\rho$ changes little ($0.913\to0.880$ and $0.878\to0.851$ for the small-edit settings), so they do not by themselves explain the correlation (\texttt{ties\_diag.py}). A program counts as failed if it raises, returns invalid output or exceeds the 60-s wall-clock limit of the evaluation worker. The limit is wall-clock time, so whether a program near it is scored depends on machine load: re-scoring the stored failures on a different machine scored 6 of 15 (Haiku neutral), 120 of 430 (Haiku rewrite), 4 of 4 (Sonnet neutral) and 0 of 55 (Sonnet rewrite), and one of 20 scored programs exceeded the limit when re-scored. Every search kept the outcome it saw; we did not re-score these programs. This is evaluation noise outside the minibatch noise we model, and it affects only programs whose evaluation takes close to 60 s.
{\sloppy\paragraph{Pre-registration and archive.} Predictions for each setting were committed to a version-control repository before its search was launched (commit timestamps are part of the archive). The searches of the BP/Haiku/neutral setting use seeds disjoint from the exploratory run, so they draw fresh proposals. The archive contains the program databases (SQLite snapshots; see below), the calibration and search trajectories, the pre-registration files, and the scripts \texttt{run2.py} (calibration, search, from-above), \texttt{predict.py} (pre-registered prediction), \texttt{evaluate\_setting.py} (comparison and fresh-sample baseline), \texttt{tau\_strata.py}, \texttt{kernel\_calib.py} and \texttt{drift\_walk\_vs\_search.py} (post-hoc diagnostics), and the scripts of the exploratory run. For the mechanism test, the hypotheses, decision rules and four-setting fit (\texttt{mech\_fit.py}) were committed before calibration of the two new settings began, and the evaluation script (\texttt{eval\_mech.py}) two minutes after it began, before any calibration or search result existed; each setting's $\hat L(\rho)$ (\texttt{predict\_mech.py}) was committed with its tilt prediction, before its search. \texttt{mech\_fit6.py} computes the post-hoc six-setting fit; \texttt{diag\_base\_law.py} (lineage depth, early-walk tilt), \texttt{h2\_power.py} and \texttt{h2\_matched.py} (selection effect), and \texttt{toy\_acf.py}, \texttt{toy\_inherit.py} and \texttt{toy\_above.py} (reversible null model) are the follow-up diagnostics. For the 300-step test, the hypotheses, the null-model predictions (\texttt{toy\_inherit.py}, \texttt{toy\_above.py}) and the evaluation script (\texttt{eval\_t300.py}) were committed before the continued runs were launched, and a power note (\texttt{t300\_power.py}) after launch and before any result. The committed script had not been executed against the full design before the results arrived. After the results, it was changed once, to skip a descriptive from-below column for a cell that was not continued (Haiku small edit has an $m=16$ search to 150 steps but none continued to 300); the decision rules and their inputs are unchanged. The follow-up diagnostics read a snapshot taken before the continued runs; the archive holds that snapshot and one taken after them, with checksums. \texttt{fig\_t300.py} draws Figure~\ref{fig:t300}; \texttt{beta\_check\_v2.py} (acceptance-rule check, Appendix~\ref{app:diag}; it replaces \texttt{beta\_check\_setting.py}, which drew minibatches with replacement), \texttt{h2\_noabove.py}, \texttt{h2\_robust.py}, \texttt{h2\_firstprop.py} and \texttt{toy\_sampling\_bias.py} (selection effect), \texttt{h1q\_uncertainty.py} (H1q with seed-blocked errors and approximate propagation of calibration and fit uncertainty), \texttt{calib\_windows.py} and \texttt{ms\_origin.py} are post-hoc checks added after review, as are, after the review of the previous version, \texttt{ms300\_sensitivity.py} (H6-300 sensitivity), \texttt{prompt\_L\_bootstrap.py} (test-5 calibration uncertainty), \texttt{ties\_diag.py} (ties and unchanged score vectors) and the run-clustered intervals of \texttt{h2\_firstprop.py}; \texttt{h2\_robust.py} regenerates its own output when run directly. For the multi-start test, the hypotheses, the twelve start programs (with their lineage depths), the seeds and the evaluation script (\texttt{eval\_multistart.py}, run on mock input before the commit) were committed before any step of the new runs was generated; \texttt{run\_multistart.py} is the driver, verified beforehand to reproduce a stored run from above with no new proposals, and \texttt{fig\_multistart.py} draws Figure~\ref{fig:multistart}. The pre-registration states that the in-band program closest to the target quality was taken for each walk and search seed. That rule was not applied exactly: for walk 9 program 2489 ($0.95463$) was used where 2539 ($0.95444$) was closest, for search seed 5 program 5306 ($0.95466$) where 5236 ($0.95451$) was closest, and for Haiku search seed 2 program 3499 ($0.95533$) where 3481 ($0.95513$) was closest; seven Sonnet walks reach the band and six were used (walk 21 was not). All starts lie in the registered band and were fixed before any run. The search-selected starts were first incumbents at (m, seed, step) $=$ (16, 0, 97), (64, 1, 105), (64, 2, 121), (64, 3, 142), (64, 4, 48), (64, 5, 90) (Sonnet) and (64, 2, 71), (16, 5, 116) (Haiku) (\texttt{ms\_origin.py}). Seeds 60 and 61 give child keys disjoint from all earlier runs; 22 of the 2{,}516 distinct Sonnet children and 6 of the 753 Haiku children are programs already in the database, returned by de-duplication rather than newly generated code. The Haiku arm ran in one segment. The Sonnet arm was stopped at about 2{,}100 of 3{,}600 children and restarted about 2.5 hours later with the same command; every stored child is written only after it is generated and evaluated, so the restart replayed the stored steps and generated no child twice. The first segment's log was overwritten by the restart (logs are now appended). Between the two segments the store and search code changed to add a prompt-optimisation domain registered separately; the bin-packing path is unchanged, and replaying all 24 trajectories with the current code reproduces them. The descriptive Haiku result was committed before the Sonnet arm finished; the rules had been committed before either arm started. A continuation of all multi-start chains to 300 steps, with two further hypotheses (whether the two groups stay apart, and where each lies relative to the prediction), was registered before it was launched (\texttt{prereg/multistart300.md}, \texttt{eval\_multistart300.py}, run on mock input before the commit); the first 151 steps of all 32 continued runs replay the stored runs. \emph{Evaluation fault and rerun.} The evaluation worker runs Python in isolated mode, which ignores user-installed packages; on the second machine the numerical library was installed only for the user, so every bin-packing child generated there (2{,}251 Sonnet and 424 Haiku children, all from this continuation) failed to evaluate and none was accepted. We found this after the first results had been computed and read. We installed the same library versions in a virtual environment used by the worker, re-scored the stored code of the affected children (all 2{,}675 now score; 20 previously scored programs per setting re-score identically, apart from the wall-clock timeouts noted under Programs generated), and reran the continuation with the same command, seeds and rules. Stored children were reused with their new scores, so a chain generates new children only where its path differs from the first run. The registered verdicts of the rerun are the same as those of the faulty run, with smaller margins ($D$: $0.0011\pm0.0005$ against $0.0020\pm0.0004$). Both the faulty and the rerun outputs are in the archive. The second-domain searches evaluate in-process and were not affected. For the prompt domain (test~5), \texttt{prereg/prompt\_domain.md} (hypotheses, rules, transferred line) was committed before calibration; the predictions (\texttt{predict.py}), $\rho$ (\texttt{rho\_prompt.py}) and the acceptance-slope check (\texttt{beta\_check\_v2.py}) after calibration and before search; and the addendum (\texttt{prereg/prompt\_domain\_addendum.md}, with \texttt{predict\_betafix.py} and the updated \texttt{eval\_prompt.py}, run on mock trajectories) during the searches, before any search output existed. \texttt{ptask.py} generates the instances and scores answers; raw task-model outputs are archived. They were written to a file named by the hash of the prompt text, so the one prompt evaluated in both settings, the shared root, has raw outputs for only one evaluation: the file reproduces the p\_small score ($0.50$), and the raw outputs of the p\_bold evaluation ($0.48$; 34 of 100 instances differ) were overwritten. Every other stored score vector, $4{,}222$ of $4{,}223$, is reproduced exactly from its raw outputs. The archive now stores raw outputs per setting and prompt id with checksums (\texttt{prompt\_raw/MANIFEST.json}), and \texttt{ptask.py} writes a new, never-overwritten file per evaluation.\par}

\section{Post-hoc diagnostics}\label{app:diag}
None of the analyses in this appendix was registered. Unless stated otherwise they read the program databases as they stood before the runs beyond 150 steps (a snapshot archived with checksums) and exclude chains started from above.

\paragraph{Noise scale and projected kernel.} $\tau$ is somewhat larger on walk pairs with higher-quality parents, but recomputing $\beta$ from walk pairs with parent quality above $0.950$ moves every prediction by at most $0.0003$ (\texttt{tau\_strata.py}). A projected-kernel predictor that fits the law of child quality given parent quality on walk pairs only, and solves the one-dimensional regime-R chain without assuming reversibility, predicts $0.9515$--$0.9576$, above the tilt in every setting (\texttt{kernel\_calib.py}).

\paragraph{Acceptance rule.} For each walk parent--child pair we simulate the paired gate (2{,}000 minibatches per $m$, each drawn without replacement), compute the acceptance log-odds $\log\{P(\hat D>0)/P(\hat D<0)\}$ and compare it with the Gaussian value at $z=\Delta\sqrt m/(\tau_{xy}\sqrt{(N-m)/(N-1)})$, for pairs with $0.05<|z|<1$; the slope ratio is the least-squares slope through zero, with a bootstrap error over pairs ($\le0.04$ throughout) (\texttt{beta\_check\_v2.py}). For $m=1,4,16,64$ the ratios are $1.73$, $1.15$, $1.02$, $1.02$ (Haiku neutral), $2.04$, $1.32$, $1.10$, $1.01$ (Haiku small edit), $1.57$, $1.03$, $0.99$, $0.98$ (Haiku rewrite), $2.11$, $1.27$, $1.06$, $1.01$ (Sonnet neutral), $1.99$, $1.22$, $1.04$, $1.01$ (Sonnet small edit) and $1.57$, $1.12$, $1.00$, $0.99$ (Sonnet rewrite). Single-instance ties are $50\%$, $72\%$, $34\%$, $69\%$, $66\%$ and $33\%$ of comparisons. The Gaussian approximation is accurate for $m\ge16$ and too flat for $m\le4$, most at $m=1$, where ties are frequent and strict acceptance rejects them in both directions. A version of this check reported in an earlier draft drew minibatch indices with replacement and omitted the finite-population factor; the conclusions are unchanged. The per-pair scale $\tau_{xy}$ is heterogeneous (exploratory run: 10th to 90th percentile $0.0031$ to $0.0078$); in the exploratory run, using the mean of $\beta_{xy}$ instead of the median moves the predictions by $+0.0005$, $+0.0007$, $+0.0007$ and $+0.0002$ for $m=1,4,16,64$. The check covers walk pairs, not the pairs a search visits.

\paragraph{Effective temperature and upper tail.} Solving $\E_{\pi_0e^{\beta f}}f=\text{observed}$ for $\beta$ gives $0.57$--$0.69$ (Haiku, neutral), $0.39$--$0.49$ (small edit), $0.31$--$0.51$ (Sonnet, neutral) and $0.25$--$0.43$ (Sonnet, small edit) times the pre-registered value for $m\le16$, and $1.17$--$1.53$ (Haiku) and $0.86$--$2.29$ (Sonnet) times with the rewrite instruction; because the tilt is flat in $\beta$ at these temperatures, the ratios are much less precise than the levels. At $m=4$ the tilt puts $51\%$, $64\%$, $66\%$ and $72\%$ of its mass above $0.953$ for the four editing settings, while $24\%$, $18\%$, $10\%$ and $36\%$ of late incumbents are above it; for the rewrite instruction the two are $30\%$ and $30\%$ (Haiku) and $26\%$ and $13\%$ (Sonnet) (\texttt{fit\_beta.py}).

\paragraph{Calibration windows.} Is the second half of the calibration walks stationary? Per setting we compare the walk mean over steps 51--75 and 76--100, paired by walk (\texttt{calib\_windows.py}). The change is $+0.0003\pm0.0009$ (Haiku neutral), $-0.0008\pm0.0003$ (Haiku small edit), $-0.0001\pm0.0008$ (Sonnet neutral), $+0.0012\pm0.0006$ (Sonnet small edit), $+0.0009\pm0.0007$ (Haiku rewrite) and $-0.0004\pm0.0005$ (Sonnet rewrite). Recomputing the $m=4$ tilt from either quarter alone moves it by at most $0.0004$, except for Haiku small edit ($0.9534$ from steps 51--75, $0.9525$ from 76--100, registered $0.9531$), where the walks are still drifting down and the choice of window changes the prediction by an amount comparable to the shortfall.

\paragraph{(a) Lineage depth.} With the rewrite instruction a child's quality barely depends on its parent's quality, but it depends on the parent's lineage depth, the length of the shortest path from the starting program in the de-duplicated proposal graph (a program reached along several paths gets the shortest; this is not necessarily the number of generations a given run passed through) (\texttt{diag\_base\_law.py}). In the Sonnet setting, children of parents at depth 1--2 average $0.9506$ and children of parents at depth 41--99 average $0.9444$. The calibration walks accept every proposal, so their second halves sit at median depth 48; the search rejects most proposals, and its incumbents sit at median depth 8 (Haiku: 39 and 8). Tilting the walk samples from steps 1--20 instead, which uses calibration data only, gives prediction errors of $-0.0003$ to $+0.0001$ for Haiku at every $m$ and $-0.0002$ to $+0.0001$ for Sonnet except at $m=4$ ($+0.0019$, the cell whose late window dips), against $-0.0008$ to $-0.0003$ and $-0.0023$ to $+0.0005$ with the registered $\pi_0$. For Haiku the dependence on depth is not monotone (children of the starting program average $0.9405$, of depth-3--5 parents $0.9480$). For the editing instructions the early walk samples are correlated with the starting program and are not a substitute for $\pi_0$. The same account predicts that the rewrite overshoot is transient: accepted steps mostly move the incumbent deeper (shortest-path depth can also fall, when a child is a shallower program already in the graph), so its level should drift down toward the deep-walk tilt. This is an untested hypothesis: no rewrite setting was continued beyond 150 steps.

\paragraph{(b) Selection effect at matched quality.} The question is whether a program that a search selected is harder to improve than a program of the same quality that a walk visited. Pooling all proposals from each parent answers it with a biased estimator. A search incumbent receives proposals until one is accepted, so the number of proposals a parent receives is larger the harder it is to improve, and pooled rates over-weight such parents; walk parents receive one proposal per visit. Weighting each parent equally biases the other way, because the last proposal of a search parent is the accepted one and more often better. In a toy model in which both kinds of parent draw their improvement probability from the same Beta$(2,6)$ law and the gate accepts a better child with probability $0.6$ and a worse one with $0.1$, the pooled rate is $0.208$ for search parents and $0.254$ for walk parents, the parent-weighted rate $0.350$, and the rate of first proposals $0.252$ (\texttt{toy\_sampling\_bias.py}). The first proposal from each parent in each run is drawn before any stopping decision and is the estimator we use. Table~\ref{tab:sel} reports the logistic regression of ``child better'' on parent quality and parent type under the three weightings, for parents in the registered band $[0.952,0.957)$ (\texttt{h2\_matched.py}, \texttt{h2\_robust.py}, \texttt{h2\_firstprop.py}). The pooled odds ratios, with intervals clustered by parent, are $0.23$--$0.53$ for the editing settings; parent weighting moves them to $0.80$--$1.45$, and first proposals give $0.51$--$1.05$, significant at the 5\% level only for Haiku small edit when clustered by parent and in no setting when clustered by run. Over the wider window $[0.948,0.957)$ search and walk parents differ in mean quality by $0.0014$--$0.0019$, more than the linear adjustment can be trusted to remove, and first-proposal odds ratios there have both signs ($0.76$, $1.41$, $0.77$, $1.26$ for the four editing settings). The earlier version of this analysis, which reported pooled odds ratios with model-based intervals, overstated the effect.

\begin{table}[h]\centering\footnotesize
\caption{Selection effect at matched quality, parents in $[0.952,0.957)$, chains from above excluded: odds ratio of a better child for search vs.\ walk parents, adjusted for parent quality (95\% intervals). Pooled: all proposals, interval clustered by parent. Parent-weighted: each proposal weighted by $1/$(proposals of its parent). First: first proposal per parent and run, interval clustered by parent (in this band every parent contributes one first proposal, so this is close to unclustered) and, in the next column, by run (search seed or calibration walk; normal approximation with few clusters). $n$: parents (first), search/walk, with the number of runs in parentheses.}\label{tab:sel}
\begin{tabular}{lccccc}\toprule
setting & pooled & parent-weighted & first proposal & first, by run & $n$ first\\\midrule
Haiku neutral & $0.53$ $[0.33,0.85]$ & $1.45$ $[0.91,2.30]$ & $1.05$ $[0.63,1.75]$ & $[0.63,1.75]$ & 471/164 (12/24)\\
Haiku small edit & $0.23$ $[0.13,0.38]$ & $0.80$ $[0.49,1.29]$ & $0.51$ $[0.28,0.92]$ & $[0.16,1.59]$ & 121/152 (6/13)\\
Sonnet neutral & $0.37$ $[0.26,0.53]$ & $0.97$ $[0.73,1.29]$ & $0.74$ $[0.54,1.01]$ & $[0.53,1.02]$ & 413/443 (6/23)\\
Sonnet small edit & $0.34$ $[0.22,0.51]$ & $0.99$ $[0.71,1.37]$ & $0.80$ $[0.54,1.19]$ & $[0.51,1.26]$ & 227/413 (6/16)\\
Haiku rewrite & $1.02$ $[0.66,1.60]$ & $1.81$ $[1.11,2.94]$ & $1.26$ $[0.63,2.52]$ & $[0.60,2.62]$ & 197/254 (6/24)\\
Sonnet rewrite & $1.02$ $[0.42,2.51]$ & $2.15$ $[0.79,5.84]$ & $0.69$ $[0.13,3.67]$ & $[0.12,3.78]$ & 108/155 (6/24)\\\bottomrule
\end{tabular}\end{table}

\paragraph{(c) A reversible null model.} In the ergodic Barker null model considered here the stationary law is the tilt, so a persistent finite-budget shortfall is a transient effect within that model. To see how large such a transient can be, we use latent quality $z=x+e$, where $x$ is inherited (an AR(1) process with variance share $s$ and persistence $r$) and $e$ is drawn afresh for every child; observed quality is $f=Q(\Phi(z))$, with $Q$ the empirical quantile function of $\pi_0$, so that a walk has marginal $\pi_0$. Fitting $s$ and $r$ to the autocorrelation of the calibration walks (lags 1--10, rank-Gaussianised) gives $s=0.94,0.94,0.62,0.90$ and $r=0.983,0.971,0.847,0.955$ for Haiku small edit, Sonnet small edit, Haiku neutral and Sonnet neutral (\texttt{toy\_acf.py}). We simulate regime R with Barker acceptance at the registered $\beta$, from the starting program (\texttt{toy\_inherit.py}, \texttt{toy\_above.py}); Barker acceptance makes the model's stationary law exactly the tilt, and with probit acceptance the level would sit slightly higher (Table~\ref{tab:beta}). Over steps 100--150 the model attains $0.72$ and $0.79$ of the lift for the two small-edit settings (observed $0.73$ and $0.62$) and $1.00$ and $0.98$ for the neutral instruction (observed $0.88$ and $0.80$). It predicts a rise of $0.15$--$0.30$ in the attained fraction from steps 50--100 to steps 100--150 in the small-edit settings, where the data are flat ($0.74\to0.73$ and $0.66\to0.62$). Its runs from above stay at the prediction ($0.9537$ and $0.9549$ at $m=4,16$ for Haiku small edit), while the observed runs from above fall to $0.9508$ and $0.9509$. With $s=0.94$ a program of the starting quality has inherited almost all of its excess, so the model has no state in which such a program is good for reasons its children lose.

\end{document}